\documentclass{article}

 \usepackage[main, final]{neurips_2025}

\usepackage[utf8]{inputenc} 
\usepackage[T1]{fontenc}    
\usepackage{hyperref}       
\usepackage{url}            
\usepackage{booktabs}       
\usepackage{amsfonts}       
\usepackage{nicefrac}       
\usepackage{microtype}      
\usepackage{xcolor}         
\usepackage{subcaption}
\usepackage{graphicx}

\title{Toward a Theory of Generalizability in LLM Mechanistic Interpretability Research}

\author{Sean Trott  \\
Department of Cognitive Science\\
University of California, San Diego\\
La Jolla, CA 92093, USA \\
\texttt{\{sttrott\}@ucsd.edu}
}

\begin{document}

\maketitle

\begin{abstract}

    Research on Large Language Models (LLMs) increasingly focuses on identifying mechanistic explanations for their behaviors, yet the field lacks clear principles for determining when (and how) findings from one model instance generalize to another. This paper addresses a fundamental epistemological challenge: given a mechanistic claim about a particular model, what justifies extrapolating this finding to other LLMs---and along which dimensions might such generalizations hold? I propose five potential \textit{axes of correspondence} along which mechanistic claims might generalize, including: functional (whether they satisfy the same functional criteria), developmental (whether they develop at similar points during pretraining), positional (whether they occupy similar absolute or relative positions), relational (whether they interact with other model components in similar ways), and configurational (whether they correspond to particular regions or structures in weight-space). To empirically validate this framework, I analyze ``1-back attention heads'' (components attending to previous tokens) across pretraining in random seeds of the Pythia models (14M, 70M, 160M, 410M). The results reveal striking consistency in the \textit{developmental trajectories} of 1-back attention across models, while positional consistency is more limited. Moreover, seeds of larger models systematically show earlier onsets, steeper slopes, and higher peaks of 1-back attention. I also address possible objections to the arguments and proposals outlined here. Finally, I conclude by arguing that progress on the generalizability of mechanistic interpretability research will consist in mapping constitutive design properties of LLMs to their emergent behaviors and mechanisms.
\end{abstract}

\section{Introduction}\label{sec:intro}

The field of \textit{mechanistic interpretability} aims to uncover the internal structures (e.g., circuits or representations) that give rise to observable behavior in Large Language Models (LLMs) and other neural network-based systems \citep{olah2020zoom, merullo2023circuit, NEURIPS2024_70e5444e}. This research has the potential to deliver novel insights about the behavior of LLMs and even help build safer, more aligned models. Yet the scientific study of LLMs has yet to establish firm \textit{epistemological foundations}: although connectionist models have of course been studied for decades \citep{elman1990finding, mcclelland1981interactive}, mechanistic interpretability of LLMs is still arguably in a ``pre-paradigmatic'' stage \citep{olah2020zoom, gurnee2023finding, olahdreams} and requires further refinement of what constitutes an explanation \citep{ayonrinde_2025_expl_virtues_ss1_2, phil_exp_ayonrinde_jaburi}. The field thus faces a number of challenges relating to how knowledge is produced and evaluated. Some of these challenges have been discussed in recent literature, e.g., accurately benchmarking model ``capabilities'' \citep{raji2021ai, ivanova2023running, saxon2024benchmarks}, but others have received only cursory treatments.

In this paper, I focus on the question of \textbf{generalizability}, specifically of \textit{mechanistic claims} about LLMs. For instance, given a particular claim about the circuits present in a particular model instance, \textbf{which aspects of this claim might generalize across model instances and which do not---and what principles can we use to guide those scientific generalizations}? I first argue that the field currently lacks a coherent theory of generalizability, and point to several potential features that might help predict whether two model instances share the same mechanisms (Section \ref{sec:definitions}). This raises the question of what it means for two circuits to be the ``same'' in the first place (Section \ref{sec:generalizing_mechanisms}). Building on recent research, I propose several potential \textit{axes of correspondence} along which mechanistic claims could plausibly generalize (Section \ref{subsec:correspondences}), and also enumerate \textit{functional criteria} that help identify which kinds of mechanisms we might expect to be robust across models (Section \ref{subsec:good_criteria}). I then validate the utility of the proposed theoretical framework in an empirical study focusing on the  \textit{positional} and \textit{developmental} properties of ``1-back attention heads'' across random seeds of models in the Pythia suite (14M, 70M, 160M, 410M) \citep{pmlr-v202-biderman23a, van2025polypythias} (Section \ref{sec:empirical_analysis}). I find striking \textit{inter-seed} developmental consistencies within each model; developmental milestones are also highly correlated \textit{across} models, albeit with some subtle differences in timing. Finally, I consider and respond to possible objections (Section \ref{sec:objections}).

\section{Samples, populations, and the generalizability problem}\label{sec:definitions}

Many scientific disciplines aim to draw general conclusions about the target of inquiry. This approach, sometimes called ``nomothetic'', contrasts with a more descriptive approach aimed at characterizing individual cases (``idiographic``) \citep{beck1953science}. In certain fields (e.g., Cognitive Science), this target of inquiry (e.g., human cognition) is too large or abstract to be observed in its entirety: consequently, researchers rely on \textit{samples} to make inferences about the underlying \textit{population of interest}. When drawing generalizations is the goal of scientific inquiry, this \textit{unbiased} sample should be representative of the population of interest.\footnote{This is not always the case: for instance, in Cognitive Science research, English speakers from Western, industrialized countries have long been overrepresented, threatening the external validity of claims made on the basis of those samples \citep{henrich2010weirdest, blasi2022over}.} 

Research on mechanistic interpretability is arguably nomothetic in nature \citep{li2015convergent, olah2020zoom, olahdreams}, i.e., the aim is to produce generalizable or ``universalizable'' claims about model behaviors or mechanisms. Of course, in practice, interpretability research is not (and cannot be) conducted on the entire ``population'' of possible LLMs; rather, it is conducted on specific \textbf{model instances}. Here, we define a ``model instance'' $m_{A, \theta, D}$ as a system with some particular architecture $A$, initialized with some particular parameter set $\theta$, and trained on some particular dataset $D$.\footnote{Note that additional axes of variation could augment this definition (e.g., particular prompts), which would further complicate the question of identifying a suitable reference class.} While some recent work has begun to explore the issue of typologizing model instances \citep{klabunde2025similarity, yax2024phylolm}, clear principles remain elusive: what would constitute a ``representative sample'' of LLMs? This makes it difficult for researchers to specify the ``population of interest'' in any particular study, or to explain the rationale behind which model instances were sampled. 

Thus, given a \textit{mechanistic claim} obtained by studying a particular model instance (or instances), what (if anything) have we learned about model instances not in the sample? For instance, if researchers identify a putative ``circuit'' in GPT-2, what other language models are likely to possess that same circuit---and on what basis could we justify this extrapolation? In principle, the answer to these questions should be informed by the factors known to influence the behaviors and mechanisms of individual model instances, such as: \textbf{architectural properties} like model size \citep{kaplan2020scaling, pmlr-v202-biderman23a, riviere2024evaluating, schrimpf2021neural} or depth \citep{mueller-linzen-2023-plant, petty2023impact}; the amount and variety of \textbf{training data} \citep{kaplan2020scaling, grieve2025sociolinguistic, DBLP:journals/corr/abs-2408-10914, chang-etal-2024-multilinguality, conneau-etal-2020-unsupervised, zhang2025the}; and the \textbf{initial parameters} of a model \citep{bencomo2025teasing, mccoy2023modeling, marinescu2024distilling, hu2025between, frankle2018lottery, belrose2024understanding}. 

Indeed, some recent work has taken exactly this approach \citep{tigges2024llm}, mapping the developmental trajectories of multiple circuits (e.g., subject-verb agreement) among models in the Pythia suite \citep{pmlr-v202-biderman23a}; one crucial finding of this work was that these trajectories were relatively \textit{aligned} across model sizes (see also \ref{subsec:correspondences}). Another potentially relevant piece of evidence comes from work on the Platonic Representation Hypothesis \citep{DBLP:journals/corr/abs-2405-07987}, which predicts that we should observe more representational convergence between larger models trained on larger volumes of data---if this hypothesis is correct, we might also expect larger, better language models to converge on more similar mechanistic solutions.

The question of which model instances will develop a particular kind of circuit is no doubt challenging. Yet this framing points to a further, even more philosophically complicated challenge: what does it even mean to assert that two different models have the ``same'' circuit? 

\section{In what ways are two circuits the ``same''?}\label{sec:generalizing_mechanisms}

Interpretability research typically involves the application of specific techniques to particular model instances, allowing researchers to determine which model components (e.g., which attention heads) perform a particular function or embed a specific concept \citep{clark2019does, wang2022interpretability, olsson2022context, merullo2023circuit, manning2020emergent, NEURIPS2024_70e5444e, park2025does, zhang2024same}. 
The result of applying such techniques might (for example) consist in a set of \textit{head indices} believed to correspond to that function, e.g., $(L3, H1), (L4, H2)$. If a researcher's goal is idiographic (i.e., characterizing a given model instance), identifying this set might be sufficient: the circuit has been mapped in a particular model. But if a researcher's goal is nomothetic (i.e., drawing generalizations about other models), they face the question of what, in particular, could plausibly be generalized across model instances.

Yet, with some exceptions \citep{binhuraib2024topoformer}, the position of a head \textit{within} a layer in the standard transformer architecture is arbitrary: even among models with the same architecture trained on the same data, there is no intrinsic reason to expect that $(L3, H_{i=1})$ should consistently perform the same function, as opposed to some other head in the same layer $(L3, H_{i \neq 1})$. Thus, what exactly do we mean when we assert that two circuits in different model instances are ``the same''? 

\subsection{Axes of Potential Correspondence}\label{subsec:correspondences}

Here, I take inspiration from research in neurophysiology: although mechanistic heterogeneity in biological networks is well-attested \citep{prinz2004similar}, researchers nonetheless strive to make generalizations about cell types and cell functions using various axes of correspondence. These include gene expression \citep{mukamel2019perspectives}, putative function \citep{moser2008place, knierim1995place, alexander2015retrosplenial}, temporal patterns \citep{riviere2022modeling, riviere2017spike}, and anatomical connectivity \citep{bates2019neuronal, haber2022learning}. Mechanistic interpretability researchers might therefore identify analogous \textit{axes of potential correspondence} between model instances:

\begin{itemize}
    \item \textbf{Function}: Intuitively, the minimal standard for asserting circuit identity across model instances is whether components in each instance meet certain functional definitions, regardless of where in each model those components are located \citep{tigges2024llm}. Here, a claim might look like: \textit{Attention heads\footnote{Note that this framework could in principle be applied to any model component at varying levels of granularity; attention heads are simply used as an illustrative example.} performing function $X$ were identified across model instances $m_1, ..., m_n$.}
    \item \textbf{Position}: One might also expect certain functions to be performed by components in similar positions across models. Here, researchers might differentiate between \textit{absolute} position (e.g., always layer $3$) and \textit{relative} position (e.g., middle layers) \citep{riviere2024evaluating, cheng2024evidence}. Here, a claim might look like: \textit{Attention heads performing function $X$ were identified at a layer depth of $0.5$ across model instances $m_1, ..., m_n$.}
    \item \textbf{Developmental}: Just as human development is associated with particular milestones \citep{murray2007infant}, some functions might plausibly emerge at similar points in the course of training, e.g., after having encountered a given number of tokens. Indeed, empirical research suggests that multiple specialized circuits begin forming at around 2B-10B tokens \citep{tigges2024llm, olsson2022context,riviere2025tacl,  van2025polypythias, jumelet2024black}; we might also expect such findings to be marked by relative discontinuities or ``phase transitions'' during development \citep{chen2023sudden, hulatent, kangaslahti2025hidden}. Here, a claim might look like: \textit{Attention heads performing function $X$ emerged after $2B$ tokens were observed across model instances $m_1, ..., m_n$.}
    \item \textbf{Relational}: Model components could also be defined in terms of how they interact with other components. For example, induction circuits consist of an ``induction head'' and a ``previous token head'' \citep{olsson2022context, 10.5555/3692070.3693925}. Similarly, \citep{zhang2024same} report analogous circuit structures across model instances trained on different languages (or different combinations of languages). Here, a claim might look like: \textit{Attention heads performing function $X$ were identified in the layer immediately following attention heads performing function $Y$ across model instances $m_1, ..., m_n$.}
    \item \textbf{Configurational}: Finally, particular functions or concepts might correspond to particular geometric configurations (e.g., in weight-space or activation-space). Here, a claim might look like: \textit{Attention heads performing function $X$ consistently occupied $Y$ region of weight-space across model instances $m_1, ..., m_n$.}
\end{itemize}

This list is not exhaustive, but rather, provides a set of initial organizing principles that help ground claims about which mechanisms might generalize across model instances (and how). For example, this framework suggests that \textbf{induction heads} might be particularly promising candidates as generalizable model components. Induction heads participate in \textit{induction circuits}, which are responsible for detecting whether a given token $t$ has appeared earlier in a sequence (e.g., position $s$), then predicting that the subsequent token will be the one that previously occurred at position $s + 1$ \citep{elhage2021mathematical, olsson2022context}. Notably, induction heads satisfy multiple of the criteria discussed above: first, heads meeting the \textit{functional} definition emerge in model instances of various sizes \citep{olsson2022context, 10.5555/3692070.3693925}; second those heads follow similar \textit{developmental} trajectories during training \citep{10.5555/3692070.3693925}; third, they (by definition) share a common \textit{relational} structure with other heads, i.e., they participate in induction circuits \citep{olsson2022context, 10.5555/3692070.3693925}; and fourth, they tend to occupy similar \textit{relative positions} across model instances \citep{olsson2022context}. 

\subsection{On finding plausible mechanistic candidates}\label{subsec:good_criteria}

Mechanistic interpretability research aimed at identifying robust, generalizable model components might also benefit from focusing on identifying plausible candidates for mechanistic functions. The relative ``success'' of induction heads in this regard points to two additional criteria that may prove useful. First, their function is closely tied to the units over which models operate (i.e., token sequences) and was not defined \textit{a priori} in terms of abstract human constructs---in this sense, they may even satisfy the definition of ``concept enrichment'' explored by \citet{ayonrinde2025bidirectional_interp}. Crucially, tracking previous sequences of tokens in the context is an intuitive solution to the problem faced by language models (predicting upcoming tokens); there is thus a clear link between induction head function and the language model training objective.

Second, while this operation is very concrete, it may be amenable to compositional abstraction \citep{olsson2022context}: in some cases, induction heads may attend not only to exact repetitions of a token but more abstract correspondences (i.e., ``types''). This could make them relevant for \textit{in-context learning}, or ICL \citep{olsson2022context, 10.5555/3692070.3693925}, which in turn suggests that they could play a useful explanatory role in higher-level accounts of LLM behavior. Although there is debate about the extent to which induction heads are directly involved in ICL \citep{yin2025which, feucht2025dual}, efforts to connect \textit{microscopic phenomena} to \textit{macroscopic behavior} \citep{olahdreams} can serve as a useful ``North Star'' for future research.

\section{1-back Attention: An Empirical Case Study}\label{sec:empirical_analysis}

If generalizability is to be a realistic ambition, then we should hope to observe some degree of robustness across \textit{minimally different} model instances, such as different random seeds of the same (or similar) architecture trained on the same data. 

The current section presents an empirical study exploring this question, focusing on \textbf{1-back attention heads}---defined as heads that direct attention from some target token to the immediately preceding token \citep{clark2019does}. From a definitional perspective, these \textbf{1-back heads} satisfy the proposed criterion of closely tracking the actual units over which models operate (i.e., token sequences; see Section \ref{subsec:good_criteria}); as with induction heads (see Section \ref{subsec:good_criteria}), tracking the immediately preceding token seems intuitively helpful for making predictions about upcoming tokens, therefore tying the putative function of these heads to the overall training objective of the model.

Because 1-back heads are likely very useful---and also quite simple in terms of their behavior---one might expect them to emerge across many model instances, including small models, making them a suitable test case for the proposed axes of correspondence (Section \ref{subsec:correspondences}). For the \textit{reference class}, I limit the analysis to (arguably) the simplest possible ``population'': different random seeds of four model architectures (Pythia-14M, Pythia-70M, Pythia-160M, Pythia-410M) trained on the same data \citep{pmlr-v202-biderman23a}. Note also that the approach adopted below focuses on characterizing the \textit{behavior} of these heads (in terms of their attention patterns), which is a necessary but not sufficient prerequisite for firmly establishing their function as 1-back heads.

This approach allows us to address three related research questions:

\begin{enumerate}
    \item[RQ1] To what extent do we observe \textit{inter-seed} and \textit{inter-model} regularities in terms of the developmental trajectories and relative position of 1-back heads? Here, we find striking \textit{developmental regularities}, consistent with prior work on other model components \citep{jumelet2024black, olsson2022context, tigges2024llm}; evidence for positional regularity is more mixed.
    \item[RQ2] What divergences do we observe across model instances, and which (if any) properties allow us to predict these divergences? Here, I find that larger models show an \textit{earlier onset} of 1-back heads than smaller models, a steeper \textit{slope} of 1-back attention over pretraining, and a higher \textit{peak} of 1-back attention.
    \item[RQ3] What predicts \textit{convergences} in developmental trajectories across model instances? Here, different seeds of the same architecture show the strongest correlation; when comparing model instances of different sizes, higher correlations were predicted by the size of each model being compared.
\end{enumerate}

A link to a GitHub repository with code and data required to reproduce these analysis can be found at \url{https://github.com/seantrott/mechinterp_generalizability}.

\subsection{Methods}

I selected the Pythia suite of auto-regressive English models \citep{pmlr-v202-biderman23a}, focusing on the nine random seeds released for Pythia-14M, 70M, 160M, and 410M \citep{van2025polypythias}. Each model was assessed at 16 training checkpoints (i.e., all available checkpoints up to and including step $1000$, followed by step $1000$, $50000$, $100000$, and $143000$). As described in \cite{pmlr-v202-biderman23a}, each model was trained on approximately 300B tokens. All models were accessed through the HuggingFace \textit{transformers} library \cite{wolf-etal-2020-transformers} and run on a 2022 Mac laptop. The Pythia models are licensed under an Apache License, Version 2.0.

Each model at each checkpoint was presented with sentences from the Natural Stories Corpus \citep{futrell2021natural}. The Natural Stories Corpus consists of 10 English-language stories, each containing approximately 1000 words. This served as a repository of naturalistic sentences with which to probe attention head behavior (note that the behavior of attention heads was remarkably consistent across different stories; see Appendix \ref{appendix:stories}). The corpus is licensed under a Creative Commons Attribution-NonCommercial-ShareAlike 4.0 International License.

The goal was to assess the developmental and positional properties of putative \textbf{1-back attention heads}. Here, ``1-back attention'' was defined as directing attention from a target token to the immediately preceding token. For each head in each model (at each checkpoint), I calculated the average 1-back attention for each sentence.\footnote{Note that qualitatively identical results for the developmental analyses (though not the positional analyses) were obtained with alternative operationalizations, such as the ratio between the average 1-back attention and the average self-attention. The two metrics were generally highly correlated within each model at the final step ($r \ge  0.68$ for all models). Average 1-back attention was favored in the final analysis because it did not depend upon arbitrary assumptions about the appropriate baseline; moreover, because the attention scores are normalized, the average 1-back attention can be interpreted as the proportion of attention directed by a given head to previous tokens.} More precisely, if $A_h(i, j)$ represents the attention assigned by head $h$ from token $i$ to token $j$, and $n$ represents the number of tokens in a given sentence I calculated:

\begin{equation}\label{equation
:ratio}
R_h = \frac{1}{n-1} \sum_{i=2}^{n} A_h(i, i-1)
\end{equation}

Note that $i$ starts at $2$ to exclude the first token in the sequence. All analyses and visualizations were conducted in R \citep{rcoreteam}.

\subsection{Results}

 \begin{figure}
    \centering
    \includegraphics[width=0.9\linewidth]{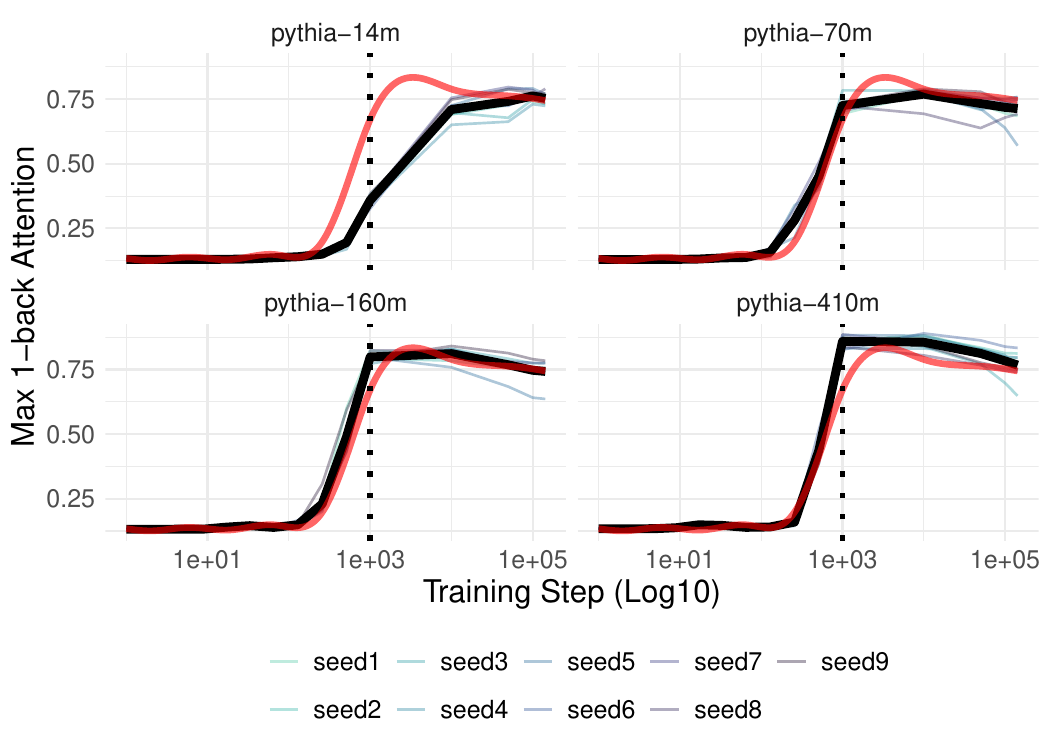}
    \caption{Maximum 1-back attention for each seed of Pythia 14M, 70M, 160M, and 410M. Dark black line indicates average across each seed for that model. Red reference line indicates predictions from a generalized additive model (GAM) fit to all data points (i.e., across models).}
    \label{fig:fig1}
\end{figure}

\subsubsection{RQ1: Inter-seed consistency}\label{results:rq1}

First, I asked about the developmental and positional consistency of putative 1-back attention across different seeds of the same architecture. As depicted in Figure \ref{fig:fig1}, different seeds of the same model showed striking regularity in their developmental trajectories. Even different models exhibited remarkably similar patterns: selective 1-back attention tended to emerge around $10^3$ training steps, corresponding to roughly $2B$ tokens of exposure. A generalized additive model (GAM) fit to all data points from all models (i.e., the maximum attention at each step for each seed of each model) using Log Training Step as a predictor achieved an $R^2 = 0.95$; as depicted in Figure \ref{fig:fig1}, the GAM's predictions (in red) are consistent with an expected onset in 1-back attention occurring between $512$ and $2000$ steps.

The \textit{position} of these putative 1-back heads exhibited considerably more variance across seeds and especially across models. Figure \ref{fig1:final_step_heatmap} depicts the maximum 1-back attention at each \textit{layer} of each random seed for each model at the final pre-training step. For $14M$, 1-back heads tended to emerge either in layer $3$ or in layer $4$\footnote{This echoes other work revealing \textit{bimodal} distributions in the mechanisms and behaviors that emerge across random seeds \citep{zhao2025distributional}, and suggests that there might be multiple ``attractors'' in weight-space with the respect to 1-back attention head emergence.}; see also Appendix \ref{appendix:ind_heads}. Overall, positional consistency across model architectures was relatively low, though there was some evidence for middle layers showing a peak in 1-back attention. A linear mixed effects model predicting average 1-back attention at the final step from each individual head of each model instance revealed a significantly negative effect of Layer Depth [$\beta = -0.11, SE = 0.005, p < 0.001$], i.e., \textit{later} layers were associated with significantly less attention to previous tokens on average. 

Together, these results point to a high degree of \textit{developmental correspondence} in putative 1-back attention heads across model instances in the Pythia suite, but also suggest a limited degree of \textit{positional correspondence}: 1-back heads are systematic in \textit{when} but not \textit{where} they appear.

\subsubsection{RQ2: Model divergences in timing}\label{results:rq2}

Although 1-back attention heads were extremely consistent in \textit{when} they developed across model instances, subtle differences in timing are revealed by comparing the trajectory of each individual model to the fit GAM predictions (Figure \ref{fig:fig1}). Relative to the predictions, smaller models (like $14M$) had a delayed \textit{onset} of 1-back heads, a shallower \textit{slope} of 1-back attention over time, and a reduced \textit{peak}; conversely, larger models (like $410M$) showed a sharper \textit{slope} and a higher \textit{peak}. 

To quantify these apparent divergences, I first operationalized each construct (onset, slope, and peak) as follows. The \textit{onset} of 1-back attention for a given seed was defined as the earliest step where the change in maximum 1-back attention relative to the change in log training step ($d_{ratio}$) exceeded some threshold; in order to avoid dependence on a particular threshold, this was assessed for a range of thresholds $(0.01, 0.3)$ for each seed and averaged across the resulting values for that range. Intuitively, this measure reflects the average \textit{earliest step} at which 1-back attention sharply increased for a given instance.\footnote{Note that the results are robust to different thresholds, as well as different operationalizations of 1-back attention onset.} The \textit{slope} was identified by regressing the maximum 1-back attention at each step against Log Training Step and extracting the resulting slope estimate. Finally, the \textit{peak} was defined as the maximum 1-back attention (i.e., across all steps for that seed). 

\begin{figure}
    \centering
    \begin{subfigure}[t]{0.48\textwidth}
        \centering
        \includegraphics[width=\linewidth]{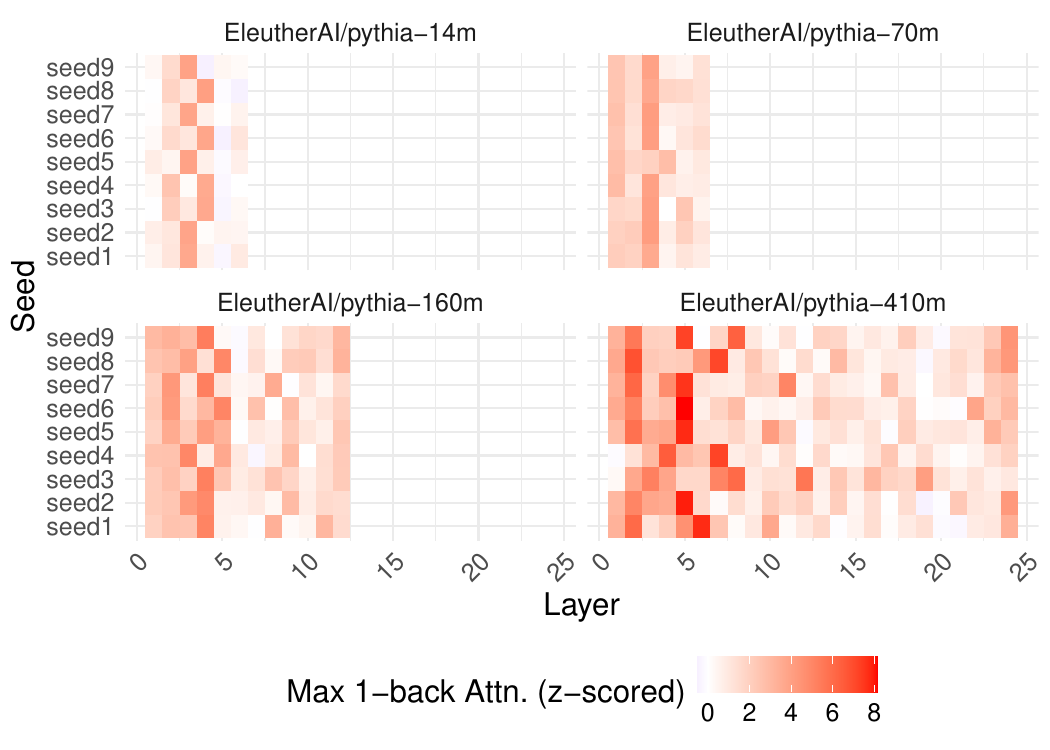}
        \caption{$Z$-scored 1-back attention for each head at the final step of each Pythia-14M random seed. Dotted red line represents $3$ standard deviations from the mean for that seed.}
        \label{fig1:final_step_heatmap}
    \end{subfigure}
    \hfill
    \begin{subfigure}[t]{0.48\textwidth}
        \centering
        \includegraphics[width=\linewidth]{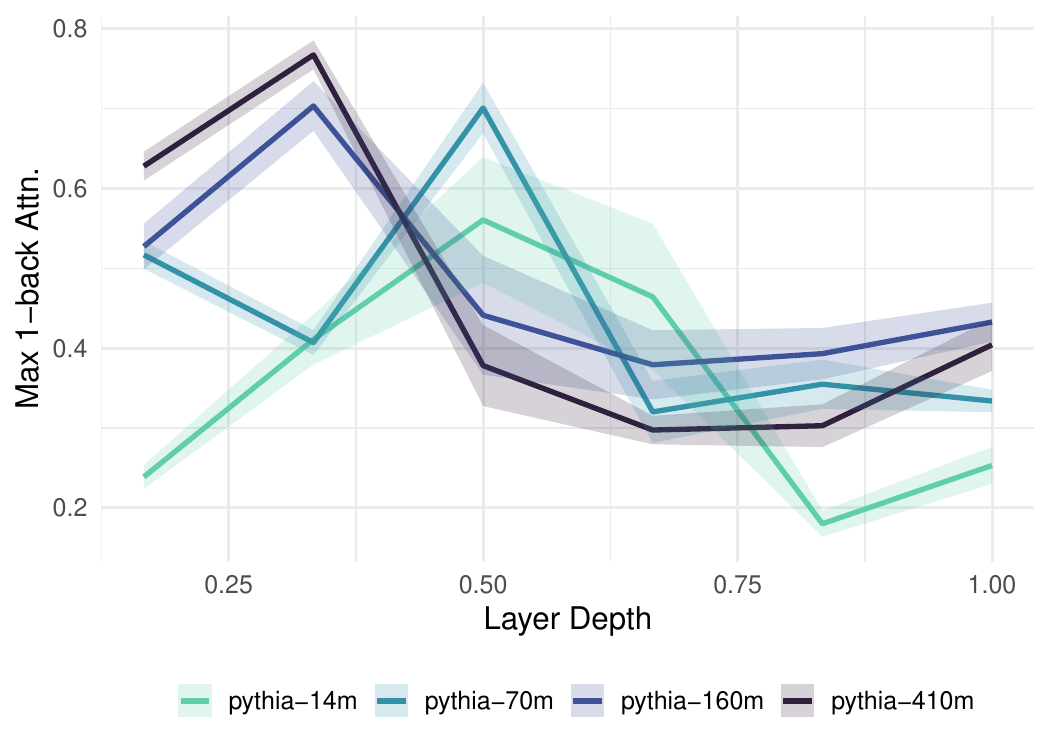}
        \caption{Maximum attention to previous tokens at each \textit{binned layer depth} across models (shading indicates one standard error calculated across seeds).}
        \label{fig4:layer_depth}
    \end{subfigure}
    \caption{Different random seeds of the same architecture showed considerable variance in where putative 1-back attention heads formed.}
    \label{fig:fig2}
\end{figure}

Each measure was then regressed (in a separate linear mixed effects model) against Log Parameters, with seed as a random intercept. Log Parameters was significantly related with each dependent variable in the expected direction. That is, larger models displayed a reduced \textit{onset} of 1-back attention $[\beta = -0.18, SE = 0.05, p <.001]$, increased \textit{slope} $[\beta = 0.01, SE = 0.003, p < .001]$, and a higher \textit{peak} $[\beta = 0.07, SE = 0.01, p < 0.001]$. These relationships are also depicted in Figure \ref{fig:fig2}. 

\subsubsection{RQ3: What predicts convergence?}\label{results:rq3}

A central question concerning generalizability is what, if anything, predicts that two model instances will belong to the same ``reference class'' with some respect to some \textit{axis of correspondence} (Section \ref{subsec:correspondences}). As described in Section \ref{results:rq1}, I observed striking convergence in the developmental trajectories of 1-back attention across random seeds of the same model, and even relatively strong temporal alignment between models of different sizes. Here, still focusing on the \textit{temporal axis}, I asked a related question: which properties predict higher temporal convergence between two model instances?

First, I calculated the Pearson's correlation between every pair of 1-back attention trajectories (i.e., for each pair of model instances). Unsurprisingly, random seeds of the same \textit{architecture} exhibited substantially higher correlation ($M = 0.99, SD = 0.04$) on average than instances from different architectures ($M = 0.95, SD = 0.003$). I then constructed a linear regression with correlation between each pair of model instances ($r_{i, j}$) as a dependent variable; predictors included a factor indicating whether the two instances were the Same Model, as well as the Number of Parameters (Log10) of model $m_i$ and model $m_j$. Consistent with the descriptive results, Same Model positively predicted higher $r_{i, j}$ $[\beta = 0.05, SE = 0.002, p <.001]$; holding Same Model constant, both Number of Parameter predictors were also positively related with higher $r_{i, j}$ $[\beta = 0.03, SE = 0.002, p <.001]$. That is, stronger temporal convergence was observed among instances belonging to the \textit{same} architecture and among instances of \textit{larger} (different) architectures; similar results are reported in the Appendix (Section \ref{appendix:mds}). 

\begin{figure}
    \centering    
    \begin{subfigure}[t]{0.32\textwidth}
        \centering
        \includegraphics[width=\linewidth]{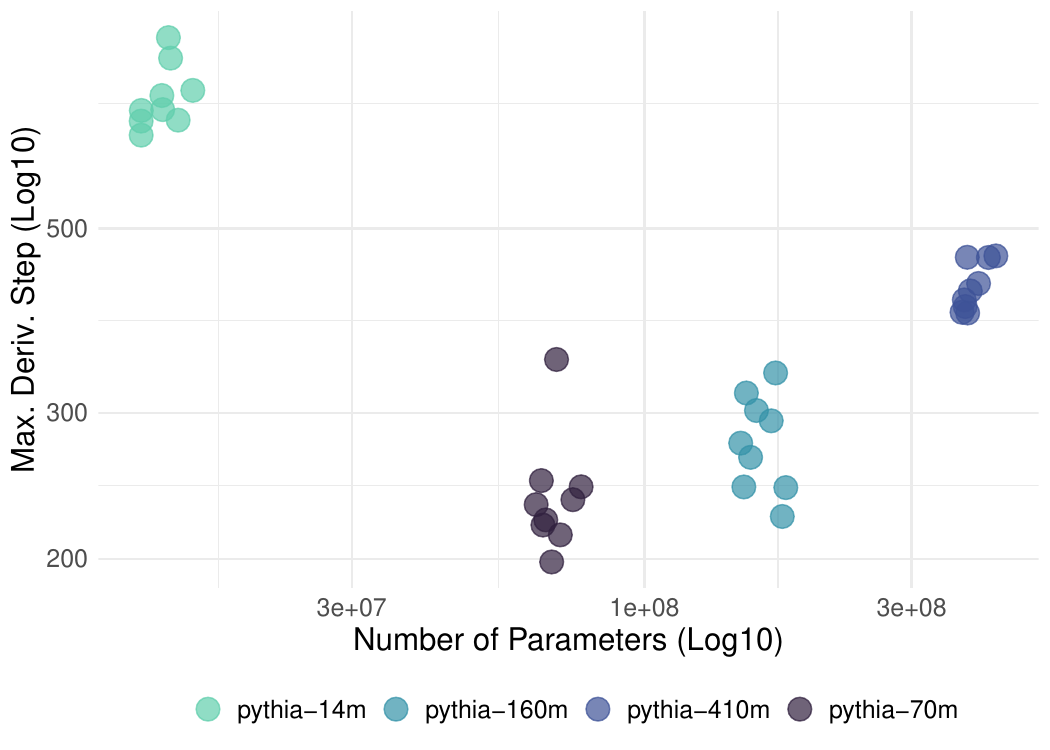}
        \caption{Earliest \textit{onset} of 1-back attention heads in each seed of each model.}
        \label{fig3:onset}
    \end{subfigure}
    \hfill
    \begin{subfigure}[t]{0.32\textwidth}
        \centering
        \includegraphics[width=\linewidth]{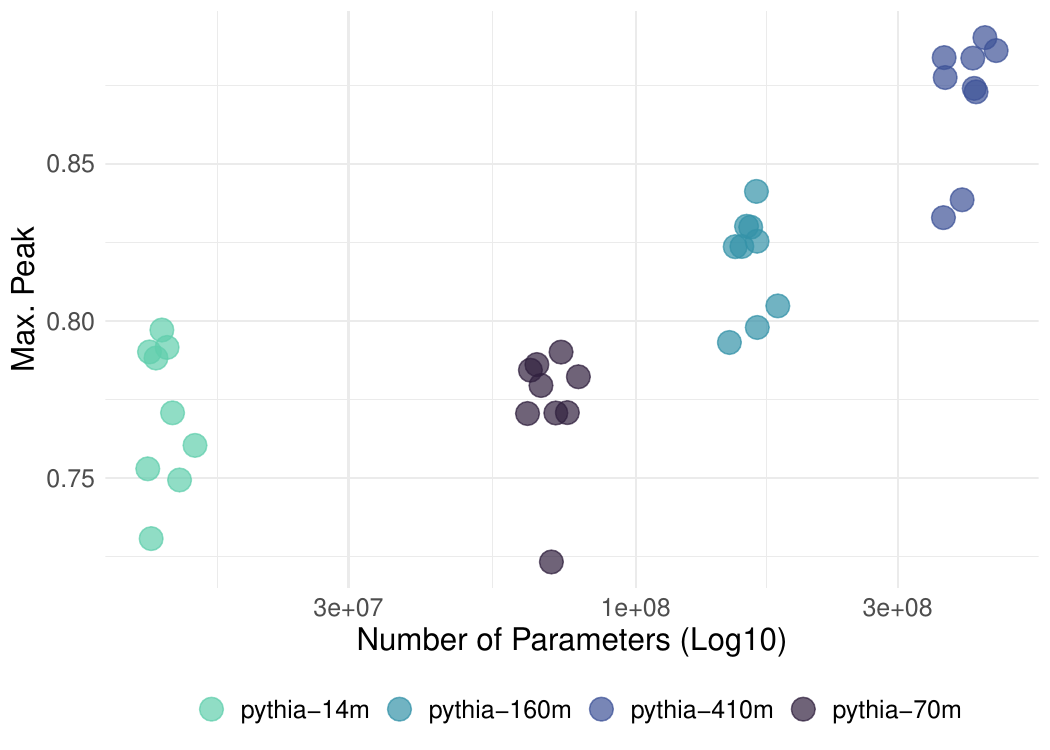}
        \caption{Max \textit{peak} of 1-back attention heads in each seed of each model.}
        \label{fig3:peak}
    \end{subfigure}
    \hfill
    \begin{subfigure}[t]{0.32\textwidth}
        \centering
        \includegraphics[width=\linewidth]{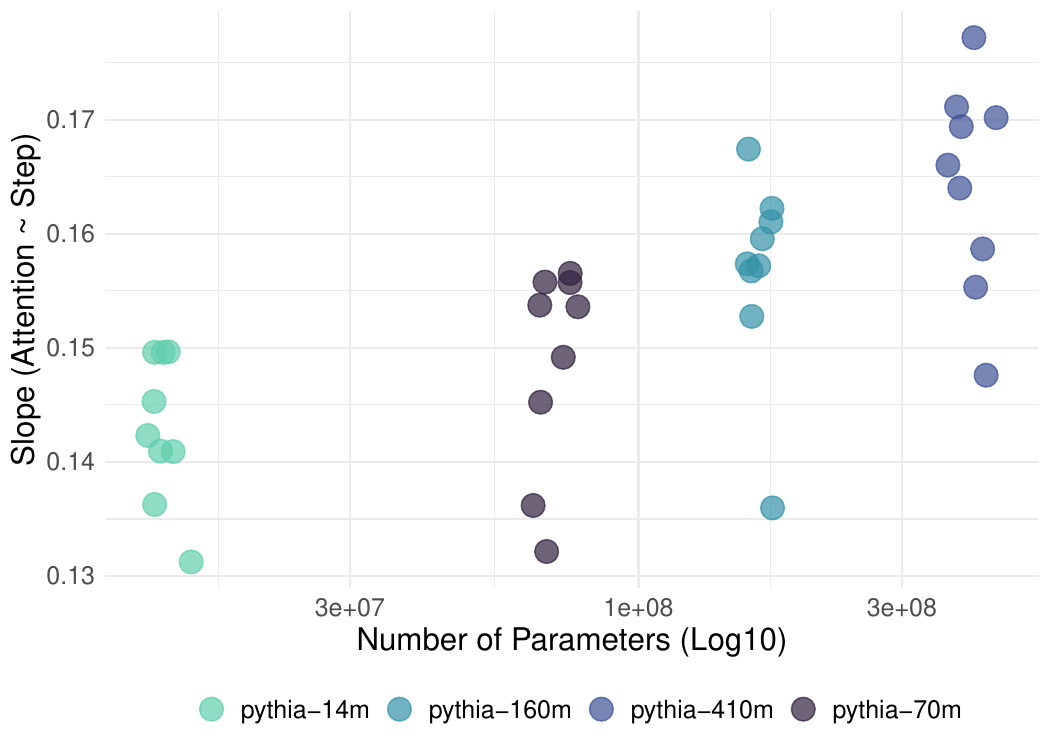}
        \caption{\textit{Slope} of 1-back attention (against pretraining step) in each seed of each model.}
        \label{fig3:slope}
    \end{subfigure}
    \hfill
    \caption{Larger Pythia models tended to show a slightly earlier \textit{onset} of 1-back attention, a higher \textit{peak}, and a steeper \textit{slope}.}
    \label{fig:fig3}
\end{figure}

\subsection{Discussion}

The primary goal of this empirical case study was as a \textit{demonstration of viability} for the proposed \textit{axes of correspondence} (Section \ref{subsec:correspondences})---specifically focusing on the positional and temporal axes of model similarity. 1-back attention heads were selected because they were a well-established phenomenon \citep{clark2019does} that would likely emerge in even small models, and which also satisfied the proposed criteria for plausible mechanistic candidates (Section \ref{subsec:good_criteria}). 

Concretely, the case study addressed three research questions. First, I observed considerable \textit{inter-seed} and \textit{inter-model} convergence along the temporal axis (see Figure \ref{fig:fig1}), though there was less positional consistency: 1-back heads were highly systematic in \textit{when} but not \textit{where} they emerged. Second, I observed subtle timing differences as a function of model size: larger models showed an earlier onset, steeper slope, and higher peak of 1-back attention (see Figure \ref{fig:fig2}). Finally, inter-seed temporal convergence was (unsurprisingly) highest among instances of the same architecture; among instances of different architectures, temporal convergence was higher when both instances were larger (Section \ref{results:rq3}), possibly consistent with the Platonic Representation Hypothesis \citep{DBLP:journals/corr/abs-2405-07987}. Together, these results suggest that at least when it comes to putative 1-back attention heads, model mechanisms are more constrained by \textit{developmental} features than \textit{positional ones}, which is itself informative about the nature of various constraints operating over head specialization.

A key limitation of this work is that candidate 1-back heads were defined in terms of their attentional behavior---their \textit{functional} role in model predictions was not investigated. Future work could conduct further analyses to investigate the functional axis specifically, perhaps connecting these heads to broader circuits in which they may or may not participate; the latter goal would also connect to other potential proposed axes of correspondence, such as the \textit{relational} structure of heads to other model components.

\section{Objections and Limitations}\label{sec:objections}

Thus far, I have argued: first, that \textit{generalizability} is a central epistemological challenge in the science of LLM interpretability; and second, that progress could be made by considering particular \textit{axes of correspondence} among model instances. I have also presented a case study illustrating the value of using these proposed axes to guide empirical investigation. Here, I consider possible objections to the key arguments in the paper, as well as potential replies to each objection. 

\textbf{Objection: Interpretability \textit{is} idiographic, not nomothetic.}

\textit{Reply:} If interpretability is idiographic, then generalizability is indeed not a concern. That said, interpretability research does (arguably) appear to be nomothetic in nature, with identifying ``universal'' circuits as a high-level goal \citep{olah2020zoom, olahdreams}. Further, as argued in \citet{olah2020zoom}, (near-)universality would also make interpretability a much more \textit{tractable} field:

\begin{quote}
    In the same way, the universality hypothesis determines what form of circuits research makes sense. If it was true in the strongest sense, one could imagine a kind of “periodic table of visual features” which we observe and catalogue across models. On the other hand, if it was mostly false, we would need to focus on a handful of models of particular societal importance and hope they stop changing every year. There might also be in between worlds, where some lessons transfer between models but others need to be learned from scratch. \citep{olah2020zoom}
\end{quote}

\textbf{Objection: Automated interpretability techniques reduce the need for generalization, as putative circuits can quickly be discovered in new model instances.}

\textit{Reply:} In a world of fully automated interpretability \citep{NEURIPS2023_34e1dbe9}, a suite of methods would simply be applied to each model instance as necessary, perhaps reducing the need to speculate about ``unobserved`` instances. However, my position is that a coherent theory of generalizability would still be of value in such a world for two reasons. First, a theory of generalizability is, at root, an articulation of what makes two model instances similar or different along some dimension; it is not only \textit{useful} for conducting research but could also be seen as a \textit{progress marker} of research. Second, in such a world there would presumably be vast empirical data characterizing large numbers of individual model instances---a theoretical framework such as the one proposed here (Section \ref{subsec:correspondences}) would give researchers a \textit{lexicon} \citep{kuhn1982commensurability,contreras2018exploratory} to describe observed convergences and divergences between model instances and make sense of this landscape.

\textbf{Objection: The case study is limited---1-back heads are simple, and a narrow reference class was chosen.}

\textit{Reply:} The goal of the case study was to investigate the viability of identifying correspondences across model instances, which is why a simple phenomenon (1-back heads) was selected, as well as a relatively narrow reference class (different seeds or sizes of the same underlying architecture trained on the same data). The underlying logic was that if generalizability is to be a goal of interpretability research, we should be able to establish it in at least this case, which perhaps reflects a ``lower bound'' of tractability. The relative success of the case study suggests these axes of correspondence may serve as fruitful guides for future work, which could expand to more complex circuits or mechanisms.

\textbf{Objection: The functional axis is sufficient to establish circuit identity.}

\textit{Reply:} The functional axis of correspondence (Section \ref{subsec:correspondences}) is arguably the \textit{minimal standard} for establishing whether components of different model instances ``do the same thing''. My argument is not that other axes are \textit{necessary}, but rather that they are additional organizing principles for assessing similarities and differences between circuits. It is informative to be able to assert that two circuits satisfy the same functional criteria but emerge at different timepoints in different models---particularly if those divergences can be related to other points of departure between those models (see Section \ref{results:rq2}). Moreover, finding correspondences along these other dimensions might also strengthen our confidence in the robustness of a particular result or in the identity of circuits across model instances: intuitively, two circuits seem more conceptually and mechanistically similar if they not only satisfy the same functional definition \citep{wang2022interpretability} but also exhibit similar \textit{developmental trajectories} (or converge along other axes of correspondence). That said, functional alignment is probably the only strictly necessary axis to assert that two circuits are implementing the same function; as in evolutionary biology, different systems often achieve the same goals or perform the same computations in different ways.

\textbf{Objection: Random seeds (and pretraining checkpoints) are not available for most models, making generalizability too difficult to investigate.}

\textit{Reply:} Indeed, the problem is even worse---the space of \textit{actual} model instances is also not itself a representative sample of the distribution of \textit{possible} models. With some exceptions \citep{pmlr-v202-biderman23a}, available models are driven by specific research or commercial interests and not necessarily with the goal of exhaustively characterizing the space of possible models. However, the fact that generalizability will be hard to investigate is not a refutation of its epistemological importance for the field of interpretability. Fully addressing the gap is beyond the scope of this paper (and will likely require large-scale coordination between institutions in multiple sectors), but individual researchers do still have options available to them, such as the Pythia suite \citep{pmlr-v202-biderman23a, van2025polypythias} and OLMo 2 \citep{olmo20252olmo2furious}. Moreover, not every interpretability study needs to address every axis of correspondence: a study investigating developmental convergences across seeds might focus on the Pythia suite \citep{pmlr-v202-biderman23a}, but a study focused on positional or relational consistency would not necessarily need pretraining checkpoints. Nonetheless, the field would clearly benefit from a larger number of  open-source models subjected to controlled training regimes.

\section{The Path Forward}\label{sec:proposals}

The shift from a pre-paradigmatic stage of research \citep{olah2020zoom, olahdreams} to more established research practices and inferential principles will require a combination of theoretical and methodological refinement. In this paper, I have argued that \textit{generalizability} is a major epistemological gap in the scientific study of LLM mechanisms. Yet enumerating challenges is often straightforward; identifying markers of progress on those challenges can be much more difficult. In that spirit, I have drawn on the growing body of existing research \citep{tigges2024llm, zhang2024same, olsson2022context} to propose \textit{axes of correspondence} that might serve as organizing principles with which to guide questions about generalizability. I have also presented the results of a case study validating the utility of this framework in identifying areas of convergence and divergence between model instances. Moving forward, one marker to look for is the construction of a theoretically legible typology (or ``phylogeny'') that makes clearly articulable predictions about which pairs of model instances will share similar mechanisms along which axes of correspondence.




\section{Reproducibility Statement}

A link to a GitHub repository with code and data required to reproduce these analyses can be found at \url{https://github.com/seantrott/mechinterp_generalizability}.

\begin{ack}
Thank you to Pamela Rivière, Kola Ayonrinde, Cameron Jones, and Oisín Parkinson-Coombs
for comments on an earlier version of this draft. The work relied on the HuggingFace \textit{transformers} package \citep{wolf-etal-2020-transformers}, the open-source Pythia suite \citep{pmlr-v202-biderman23a, van2025polypythias}, the Natural Stories Corpus \citep{futrell2021natural}, and the open-source R computing environment \citep{rcoreteam}.
\end{ack}


\bibliographystyle{plainnat}
\bibliography{neuerips}

\begin{thebibliography}{75}
\providecommand{\natexlab}[1]{#1}
\providecommand{\url}[1]{\texttt{#1}}
\expandafter\ifx\csname urlstyle\endcsname\relax
  \providecommand{\doi}[1]{doi: #1}\else
  \providecommand{\doi}{doi: \begingroup \urlstyle{rm}\Url}\fi

\bibitem[Alexander and Nitz(2015)]{alexander2015retrosplenial}
Andrew~S Alexander and Douglas~A Nitz.
\newblock Retrosplenial cortex maps the conjunction of internal and external spaces.
\newblock \emph{Nature neuroscience}, 18\penalty0 (8):\penalty0 1143--1151, 2015.

\bibitem[Aryabumi et~al.(2024)Aryabumi, Su, Ma, Morisot, Zhang, Locatelli, Fadaee, Üstün, and Hooker]{DBLP:journals/corr/abs-2408-10914}
Viraat Aryabumi, Yixuan Su, Raymond Ma, Adrien Morisot, Ivan Zhang, Acyr Locatelli, Marzieh Fadaee, Ahmet Üstün, and Sara Hooker.
\newblock To code, or not to code? exploring impact of code in pre-training.
\newblock \emph{CoRR}, abs/2408.10914, 2024.
\newblock URL \url{https://doi.org/10.48550/arXiv.2408.10914}.

\bibitem[Ayonrinde(2025)]{ayonrinde2025bidirectional_interp}
Kola Ayonrinde.
\newblock Position: Interpretability is a bidirectional communication problem.
\newblock In \emph{ICLR 2025 Workshop on Bidirectional Human-AI Alignment}, 2025.
\newblock URL \url{https://openreview.net/forum?id=O4LaRH4zSI}.

\bibitem[Ayonrinde and Jaburi(2025{\natexlab{a}})]{ayonrinde_2025_expl_virtues_ss1_2}
Kola Ayonrinde and Louis Jaburi.
\newblock Evaluating explanations: An explanatory virtues framework for mechanistic interpretability.
\newblock 2025{\natexlab{a}}.
\newblock Forthcoming.

\bibitem[Ayonrinde and Jaburi(2025{\natexlab{b}})]{phil_exp_ayonrinde_jaburi}
Kola Ayonrinde and Louis Jaburi.
\newblock A mathematical philosophy of explanations in mechanistic interpretability: The strange science part i.i, 2025{\natexlab{b}}.
\newblock forthcoming.

\bibitem[Bates et~al.(2019)Bates, Janssens, Jefferis, and Aerts]{bates2019neuronal}
Alexander~Shakeel Bates, Jasper Janssens, Gregory~Sxe Jefferis, and Stein Aerts.
\newblock Neuronal cell types in the fly: single-cell anatomy meets single-cell genomics.
\newblock \emph{Current opinion in neurobiology}, 56:\penalty0 125--134, 2019.

\bibitem[Beck(1953)]{beck1953science}
Samuel~J Beck.
\newblock The science of personality: Nomothetic or idiographic?
\newblock \emph{Psychological Review}, 60\penalty0 (6):\penalty0 353, 1953.

\bibitem[Belrose and Scherlis(2024)]{belrose2024understanding}
Nora Belrose and Adam Scherlis.
\newblock Understanding gradient descent through the training jacobian.
\newblock \emph{arXiv preprint arXiv:2412.07003}, 2024.

\bibitem[Bencomo et~al.(2025)Bencomo, Gupta, Marinescu, McCoy, and Griffiths]{bencomo2025teasing}
Gianluca Bencomo, Max Gupta, Ioana Marinescu, R~Thomas McCoy, and Thomas~L Griffiths.
\newblock Teasing apart architecture and initial weights as sources of inductive bias in neural networks.
\newblock \emph{arXiv preprint arXiv:2502.20237}, 2025.

\bibitem[Biderman et~al.(2023)Biderman, Schoelkopf, Anthony, Bradley, O'Brien, Hallahan, Khan, Purohit, Prashanth, Raff, Skowron, Sutawika, and Van Der~Wal]{pmlr-v202-biderman23a}
Stella Biderman, Hailey Schoelkopf, Quentin~Gregory Anthony, Herbie Bradley, Kyle O'Brien, Eric Hallahan, Mohammad~Aflah Khan, Shivanshu Purohit, Usvsn~Sai Prashanth, Edward Raff, Aviya Skowron, Lintang Sutawika, and Oskar Van Der~Wal.
\newblock Pythia: A suite for analyzing large language models across training and scaling.
\newblock In Andreas Krause, Emma Brunskill, Kyunghyun Cho, Barbara Engelhardt, Sivan Sabato, and Jonathan Scarlett, editors, \emph{Proceedings of the 40th International Conference on Machine Learning}, volume 202 of \emph{Proceedings of Machine Learning Research}, pages 2397--2430. PMLR, 23--29 Jul 2023.
\newblock URL \url{https://proceedings.mlr.press/v202/biderman23a.html}.

\bibitem[Binhuraib et~al.(2024)Binhuraib, Tuckute, and Blauch]{binhuraib2024topoformer}
Taha Osama~A Binhuraib, Greta Tuckute, and Nicholas Blauch.
\newblock Topoformer: brain-like topographic organization in transformer language models through spatial querying and reweighting.
\newblock In \emph{ICLR 2024 Workshop on Representational Alignment}, 2024.

\bibitem[Blasi et~al.(2022)Blasi, Henrich, Adamou, Kemmerer, and Majid]{blasi2022over}
Dami{\'a}n~E Blasi, Joseph Henrich, Evangelia Adamou, David Kemmerer, and Asifa Majid.
\newblock Over-reliance on english hinders cognitive science.
\newblock \emph{Trends in cognitive sciences}, 26\penalty0 (12):\penalty0 1153--1170, 2022.

\bibitem[Carroll and Arabie(1998)]{carroll1998multidimensional}
J~Douglas Carroll and Phipps Arabie.
\newblock Multidimensional scaling.
\newblock \emph{Measurement, judgment and decision making}, pages 179--250, 1998.

\bibitem[Chang et~al.(2024)Chang, Arnett, Tu, and Bergen]{chang-etal-2024-multilinguality}
Tyler~A. Chang, Catherine Arnett, Zhuowen Tu, and Ben Bergen.
\newblock When is multilinguality a curse? language modeling for 250 high- and low-resource languages.
\newblock In Yaser Al-Onaizan, Mohit Bansal, and Yun-Nung Chen, editors, \emph{Proceedings of the 2024 Conference on Empirical Methods in Natural Language Processing}, pages 4074--4096, Miami, Florida, USA, November 2024. Association for Computational Linguistics.
\newblock \doi{10.18653/v1/2024.emnlp-main.236}.
\newblock URL \url{https://aclanthology.org/2024.emnlp-main.236/}.

\bibitem[Chen et~al.(2023)Chen, Shwartz-Ziv, Cho, Leavitt, and Saphra]{chen2023sudden}
Angelica Chen, Ravid Shwartz-Ziv, Kyunghyun Cho, Matthew~L Leavitt, and Naomi Saphra.
\newblock Sudden drops in the loss: Syntax acquisition, phase transitions, and simplicity bias in mlms.
\newblock \emph{arXiv preprint arXiv:2309.07311}, 2023.

\bibitem[Cheng and Antonello(2024)]{cheng2024evidence}
Emily Cheng and Richard~J Antonello.
\newblock Evidence from fmri supports a two-phase abstraction process in language models.
\newblock \emph{arXiv preprint arXiv:2409.05771}, 2024.

\bibitem[Clark et~al.(2019)Clark, Khandelwal, Levy, and Manning]{clark2019does}
Kevin Clark, Urvashi Khandelwal, Omer Levy, and Christopher~D Manning.
\newblock What does bert look at? an analysis of bert’s attention.
\newblock In \emph{Proceedings of the 2019 ACL Workshop BlackboxNLP: Analyzing and Interpreting Neural Networks for NLP}, page 276. Association for Computational Linguistics, 2019.

\bibitem[Conmy et~al.(2023)Conmy, Mavor-Parker, Lynch, Heimersheim, and Garriga-Alonso]{NEURIPS2023_34e1dbe9}
Arthur Conmy, Augustine Mavor-Parker, Aengus Lynch, Stefan Heimersheim, and Adri\`{a} Garriga-Alonso.
\newblock Towards automated circuit discovery for mechanistic interpretability.
\newblock In A.~Oh, T.~Naumann, A.~Globerson, K.~Saenko, M.~Hardt, and S.~Levine, editors, \emph{Advances in Neural Information Processing Systems}, volume~36, pages 16318--16352. Curran Associates, Inc., 2023.
\newblock URL \url{https://proceedings.neurips.cc/paper_files/paper/2023/file/34e1dbe95d34d7ebaf99b9bcaeb5b2be-Paper-Conference.pdf}.

\bibitem[Conneau et~al.(2020)Conneau, Khandelwal, Goyal, Chaudhary, Wenzek, Guzm{\'a}n, Grave, Ott, Zettlemoyer, and Stoyanov]{conneau-etal-2020-unsupervised}
Alexis Conneau, Kartikay Khandelwal, Naman Goyal, Vishrav Chaudhary, Guillaume Wenzek, Francisco Guzm{\'a}n, Edouard Grave, Myle Ott, Luke Zettlemoyer, and Veselin Stoyanov.
\newblock Unsupervised cross-lingual representation learning at scale.
\newblock In Dan Jurafsky, Joyce Chai, Natalie Schluter, and Joel Tetreault, editors, \emph{Proceedings of the 58th Annual Meeting of the Association for Computational Linguistics}, pages 8440--8451, Online, July 2020. Association for Computational Linguistics.
\newblock \doi{10.18653/v1/2020.acl-main.747}.
\newblock URL \url{https://aclanthology.org/2020.acl-main.747/}.

\bibitem[Contreras~Kallens and Dale(2018)]{contreras2018exploratory}
Pablo Contreras~Kallens and Rick Dale.
\newblock Exploratory mapping of theoretical landscapes through word use in abstracts.
\newblock \emph{Scientometrics}, 116\penalty0 (3):\penalty0 1641--1674, 2018.

\bibitem[Elhage et~al.(2021)Elhage, Nanda, Olsson, Henighan, Joseph, Mann, Askell, Bai, Chen, Conerly, et~al.]{elhage2021mathematical}
Nelson Elhage, Neel Nanda, Catherine Olsson, Tom Henighan, Nicholas Joseph, Ben Mann, Amanda Askell, Yuntao Bai, Anna Chen, Tom Conerly, et~al.
\newblock A mathematical framework for transformer circuits.
\newblock \emph{Transformer Circuits Thread}, 1\penalty0 (1):\penalty0 12, 2021.

\bibitem[Elman(1990)]{elman1990finding}
Jeffrey~L Elman.
\newblock Finding structure in time.
\newblock \emph{Cognitive science}, 14\penalty0 (2):\penalty0 179--211, 1990.

\bibitem[Feucht et~al.(2025)Feucht, Todd, Wallace, and Bau]{feucht2025dual}
Sheridan Feucht, Eric Todd, Byron Wallace, and David Bau.
\newblock The dual-route model of induction.
\newblock \emph{arXiv preprint arXiv:2504.03022}, 2025.

\bibitem[Frankle and Carbin(2018)]{frankle2018lottery}
Jonathan Frankle and Michael Carbin.
\newblock The lottery ticket hypothesis: Finding sparse, trainable neural networks.
\newblock In \emph{International Conference on Learning Representations}, 2018.

\bibitem[Futrell et~al.(2021)Futrell, Gibson, Tily, Blank, Vishnevetsky, Piantadosi, and Fedorenko]{futrell2021natural}
Richard Futrell, Edward Gibson, Harry~J Tily, Idan Blank, Anastasia Vishnevetsky, Steven~T Piantadosi, and Evelina Fedorenko.
\newblock The natural stories corpus: a reading-time corpus of english texts containing rare syntactic constructions.
\newblock \emph{Language Resources and Evaluation}, 55:\penalty0 63--77, 2021.

\bibitem[Grieve et~al.(2025)Grieve, Bartl, Fuoli, Grafmiller, Huang, Jawerbaum, Murakami, Perlman, Roemling, and Winter]{grieve2025sociolinguistic}
Jack Grieve, Sara Bartl, Matteo Fuoli, Jason Grafmiller, Weihang Huang, Alejandro Jawerbaum, Akira Murakami, Marcus Perlman, Dana Roemling, and Bodo Winter.
\newblock The sociolinguistic foundations of language modeling.
\newblock \emph{Frontiers in Artificial Intelligence}, 7:\penalty0 1472411, 2025.

\bibitem[Gurnee et~al.(2023)Gurnee, Nanda, Pauly, Harvey, Troitskii, and Bertsimas]{gurnee2023finding}
Wes Gurnee, Neel Nanda, Matthew Pauly, Katherine Harvey, Dmitrii Troitskii, and Dimitris Bertsimas.
\newblock Finding neurons in a haystack: Case studies with sparse probing.
\newblock \emph{arXiv preprint arXiv:2305.01610}, 2023.

\bibitem[Haber and Schneidman(2022)]{haber2022learning}
Adam Haber and Elad Schneidman.
\newblock Learning the architectural features that predict functional similarity of neural networks.
\newblock \emph{Physical Review X}, 12\penalty0 (2):\penalty0 021051, 2022.

\bibitem[Henrich et~al.(2010)Henrich, Heine, and Norenzayan]{henrich2010weirdest}
Joseph Henrich, Steven~J Heine, and Ara Norenzayan.
\newblock The weirdest people in the world?
\newblock \emph{Behavioral and brain sciences}, 33\penalty0 (2-3):\penalty0 61--83, 2010.

\bibitem[Hu et~al.()Hu, Chen, Saphra, and Cho]{hulatent}
Michael~Y Hu, Angelica Chen, Naomi Saphra, and Kyunghyun Cho.
\newblock Latent state models of training dynamics.
\newblock \emph{Transactions on Machine Learning Research}.

\bibitem[Hu et~al.(2025)Hu, Petty, Shi, Merrill, and Linzen]{hu2025between}
Michael~Y Hu, Jackson Petty, Chuan Shi, William Merrill, and Tal Linzen.
\newblock Between circuits and chomsky: Pre-pretraining on formal languages imparts linguistic biases.
\newblock \emph{arXiv preprint arXiv:2502.19249}, 2025.

\bibitem[Huh et~al.(2024)Huh, Cheung, Wang, and Isola]{DBLP:journals/corr/abs-2405-07987}
Minyoung Huh, Brian Cheung, Tongzhou Wang, and Phillip Isola.
\newblock The platonic representation hypothesis.
\newblock \emph{CoRR}, abs/2405.07987, 2024.
\newblock URL \url{https://doi.org/10.48550/arXiv.2405.07987}.

\bibitem[Ivanova(2023)]{ivanova2023running}
Anna~A Ivanova.
\newblock Running cognitive evaluations on large language models: The do's and the don'ts.
\newblock \emph{arXiv preprint arXiv:2312.01276}, 2023.

\bibitem[Jumelet et~al.(2024)Jumelet, Bylinina, Zuidema, and Szymanik]{jumelet2024black}
Jaap Jumelet, Lisa Bylinina, Willem Zuidema, and Jakub Szymanik.
\newblock Black big boxes: Do language models hide a theory of adjective order?
\newblock \emph{arXiv preprint arXiv:2407.02136}, 2024.

\bibitem[Kangaslahti et~al.(2025)Kangaslahti, Rosenfeld, and Saphra]{kangaslahti2025hidden}
Sara Kangaslahti, Elan Rosenfeld, and Naomi Saphra.
\newblock Hidden breakthroughs in language model training.
\newblock \emph{arXiv preprint arXiv:2506.15872}, 2025.

\bibitem[Kaplan et~al.(2020)Kaplan, McCandlish, Henighan, Brown, Chess, Child, Gray, Radford, Wu, and Amodei]{kaplan2020scaling}
Jared Kaplan, Sam McCandlish, Tom Henighan, Tom~B Brown, Benjamin Chess, Rewon Child, Scott Gray, Alec Radford, Jeffrey Wu, and Dario Amodei.
\newblock Scaling laws for neural language models.
\newblock \emph{arXiv preprint arXiv:2001.08361}, 2020.

\bibitem[Klabunde et~al.(2025)Klabunde, Schumacher, Strohmaier, and Lemmerich]{klabunde2025similarity}
Max Klabunde, Tobias Schumacher, Markus Strohmaier, and Florian Lemmerich.
\newblock Similarity of neural network models: A survey of functional and representational measures.
\newblock \emph{ACM Computing Surveys}, 57\penalty0 (9):\penalty0 1--52, 2025.

\bibitem[Knierim et~al.(1995)Knierim, Kudrimoti, and McNaughton]{knierim1995place}
James~J Knierim, Hemant~S Kudrimoti, and Bruce~L McNaughton.
\newblock Place cells, head direction cells, and the learning of landmark stability.
\newblock \emph{Journal of Neuroscience}, 15\penalty0 (3):\penalty0 1648--1659, 1995.

\bibitem[Kuhn(1982)]{kuhn1982commensurability}
Thomas~S Kuhn.
\newblock Commensurability, comparability, communicability.
\newblock In \emph{PSA: Proceedings of the biennial meeting of the Philosophy of Science Association}, volume 1982, pages 668--688. Cambridge University Press, 1982.

\bibitem[Li et~al.(2015)Li, Yosinski, Clune, Lipson, and Hopcroft]{li2015convergent}
Yixuan Li, Jason Yosinski, Jeff Clune, Hod Lipson, and John Hopcroft.
\newblock Convergent learning: Do different neural networks learn the same representations?
\newblock In \emph{Feature Extraction: Modern Questions and Challenges}, pages 196--212. PMLR, 2015.

\bibitem[Manning et~al.(2020)Manning, Clark, Hewitt, Khandelwal, and Levy]{manning2020emergent}
Christopher~D Manning, Kevin Clark, John Hewitt, Urvashi Khandelwal, and Omer Levy.
\newblock Emergent linguistic structure in artificial neural networks trained by self-supervision.
\newblock \emph{Proceedings of the National Academy of Sciences}, 117\penalty0 (48):\penalty0 30046--30054, 2020.

\bibitem[Marinescu et~al.(2024)Marinescu, McCoy, and Griffiths]{marinescu2024distilling}
Ioana Marinescu, R~Thomas McCoy, and Tom Griffiths.
\newblock Distilling symbolic priors for concept learning into neural networks.
\newblock In \emph{Proceedings of the Annual Meeting of the Cognitive Science Society}, volume~46, 2024.

\bibitem[McClelland and Rumelhart(1981)]{mcclelland1981interactive}
James~L McClelland and David~E Rumelhart.
\newblock An interactive activation model of context effects in letter perception: I. an account of basic findings.
\newblock \emph{Psychological review}, 88\penalty0 (5):\penalty0 375, 1981.

\bibitem[McCoy and Griffiths(2023)]{mccoy2023modeling}
R~Thomas McCoy and Thomas~L Griffiths.
\newblock Modeling rapid language learning by distilling bayesian priors into artificial neural networks.
\newblock \emph{arXiv preprint arXiv:2305.14701}, 2023.

\bibitem[Merullo et~al.(2023)Merullo, Eickhoff, and Pavlick]{merullo2023circuit}
Jack Merullo, Carsten Eickhoff, and Ellie Pavlick.
\newblock Circuit component reuse across tasks in transformer language models.
\newblock \emph{arXiv preprint arXiv:2310.08744}, 2023.

\bibitem[Merullo et~al.(2024)Merullo, Eickhoff, and Pavlick]{NEURIPS2024_70e5444e}
Jack Merullo, Carsten Eickhoff, and Ellie Pavlick.
\newblock Talking heads: Understanding inter-layer communication in transformer language models.
\newblock In A.~Globerson, L.~Mackey, D.~Belgrave, A.~Fan, U.~Paquet, J.~Tomczak, and C.~Zhang, editors, \emph{Advances in Neural Information Processing Systems}, volume~37, pages 61372--61418. Curran Associates, Inc., 2024.
\newblock URL \url{https://proceedings.neurips.cc/paper_files/paper/2024/file/70e5444e5f331f7f5431f302110b97af-Paper-Conference.pdf}.

\bibitem[Moser et~al.(2008)Moser, Kropff, and Moser]{moser2008place}
Edvard~I Moser, Emilio Kropff, and May-Britt Moser.
\newblock Place cells, grid cells, and the brain's spatial representation system.
\newblock \emph{Annu. Rev. Neurosci.}, 31\penalty0 (1):\penalty0 69--89, 2008.

\bibitem[Mueller and Linzen(2023)]{mueller-linzen-2023-plant}
Aaron Mueller and Tal Linzen.
\newblock How to plant trees in language models: Data and architectural effects on the emergence of syntactic inductive biases.
\newblock In Anna Rogers, Jordan Boyd-Graber, and Naoaki Okazaki, editors, \emph{Proceedings of the 61st Annual Meeting of the Association for Computational Linguistics (Volume 1: Long Papers)}, pages 11237--11252, Toronto, Canada, July 2023. Association for Computational Linguistics.
\newblock \doi{10.18653/v1/2023.acl-long.629}.
\newblock URL \url{https://aclanthology.org/2023.acl-long.629/}.

\bibitem[Mukamel and Ngai(2019)]{mukamel2019perspectives}
Eran~A Mukamel and John Ngai.
\newblock Perspectives on defining cell types in the brain.
\newblock \emph{Current opinion in neurobiology}, 56:\penalty0 61--68, 2019.

\bibitem[Murray et~al.(2007)Murray, Jones, Kuh, and Richards]{murray2007infant}
Graham~K Murray, Peter~B Jones, Diana Kuh, and Marcus Richards.
\newblock Infant developmental milestones and subsequent cognitive function.
\newblock \emph{Annals of neurology}, 62\penalty0 (2):\penalty0 128--136, 2007.

\bibitem[Olah(2023)]{olahdreams}
Chris Olah.
\newblock Interpretability dreams.
\newblock \emph{Transformer Circuits}, 2023.
\newblock https://transformer-circuits.pub/2023/interpretability-dreams/index.html.

\bibitem[Olah et~al.(2020)Olah, Cammarata, Schubert, Goh, Petrov, and Carter]{olah2020zoom}
Chris Olah, Nick Cammarata, Ludwig Schubert, Gabriel Goh, Michael Petrov, and Shan Carter.
\newblock Zoom in: An introduction to circuits.
\newblock \emph{Distill}, 2020.
\newblock \doi{10.23915/distill.00024.001}.
\newblock https://distill.pub/2020/circuits/zoom-in.

\bibitem[OLMo et~al.(2025)OLMo, Walsh, Soldaini, Groeneveld, Lo, Arora, Bhagia, Gu, Huang, Jordan, Lambert, Schwenk, Tafjord, Anderson, Atkinson, Brahman, Clark, Dasigi, Dziri, Guerquin, Ivison, Koh, Liu, Malik, Merrill, Miranda, Morrison, Murray, Nam, Pyatkin, Rangapur, Schmitz, Skjonsberg, Wadden, Wilhelm, Wilson, Zettlemoyer, Farhadi, Smith, and Hajishirzi]{olmo20252olmo2furious}
Team OLMo, Pete Walsh, Luca Soldaini, Dirk Groeneveld, Kyle Lo, Shane Arora, Akshita Bhagia, Yuling Gu, Shengyi Huang, Matt Jordan, Nathan Lambert, Dustin Schwenk, Oyvind Tafjord, Taira Anderson, David Atkinson, Faeze Brahman, Christopher Clark, Pradeep Dasigi, Nouha Dziri, Michal Guerquin, Hamish Ivison, Pang~Wei Koh, Jiacheng Liu, Saumya Malik, William Merrill, Lester James~V. Miranda, Jacob Morrison, Tyler Murray, Crystal Nam, Valentina Pyatkin, Aman Rangapur, Michael Schmitz, Sam Skjonsberg, David Wadden, Christopher Wilhelm, Michael Wilson, Luke Zettlemoyer, Ali Farhadi, Noah~A. Smith, and Hannaneh Hajishirzi.
\newblock 2 olmo 2 furious, 2025.
\newblock URL \url{https://arxiv.org/abs/2501.00656}.

\bibitem[Olsson et~al.(2022)Olsson, Elhage, Nanda, Joseph, DasSarma, Henighan, Mann, Askell, Bai, Chen, et~al.]{olsson2022context}
Catherine Olsson, Nelson Elhage, Neel Nanda, Nicholas Joseph, Nova DasSarma, Tom Henighan, Ben Mann, Amanda Askell, Yuntao Bai, Anna Chen, et~al.
\newblock In-context learning and induction heads.
\newblock \emph{arXiv preprint arXiv:2209.11895}, 2022.

\bibitem[Park et~al.(2025)Park, Yoon, Park, Jeong, and Kang]{park2025does}
Yein Park, Chanwoong Yoon, Jungwoo Park, Minbyul Jeong, and Jaewoo Kang.
\newblock Does time have its place? temporal heads: Where language models recall time-specific information.
\newblock \emph{arXiv preprint arXiv:2502.14258}, 2025.

\bibitem[Petty et~al.(2023)Petty, van Steenkiste, Sha, Dasgupta, Garrette, and Linzen]{petty2023impact}
Jackson Petty, Sjoerd van Steenkiste, Fei Sha, Ishita Dasgupta, Dan Garrette, and Tal Linzen.
\newblock The impact of depth and width on transformer language model generalization.
\newblock 2023.

\bibitem[Prinz et~al.(2004)Prinz, Bucher, and Marder]{prinz2004similar}
Astrid~A Prinz, Dirk Bucher, and Eve Marder.
\newblock Similar network activity from disparate circuit parameters.
\newblock \emph{Nature neuroscience}, 7\penalty0 (12):\penalty0 1345--1352, 2004.

\bibitem[{R Core Team}(2025)]{rcoreteam}
{R Core Team}.
\newblock \emph{R: A Language and Environment for Statistical Computing}.
\newblock R Foundation for Statistical Computing, Vienna, Austria, 2025.
\newblock URL \url{https://www.R-project.org/}.

\bibitem[Raji et~al.(2021)Raji, Denton, Bender, Hanna, and Paullada]{raji2021ai}
Inioluwa~Deborah Raji, Emily Denton, Emily~M. Bender, Alex Hanna, and Amandalynne Paullada.
\newblock {AI} and the everything in the whole wide world benchmark.
\newblock In \emph{Thirty-fifth Conference on Neural Information Processing Systems Datasets and Benchmarks Track (Round 2)}, 2021.
\newblock URL \url{https://openreview.net/forum?id=j6NxpQbREA1}.

\bibitem[Rivi{\`e}re and Rangel(2017)]{riviere2017spike}
Pamela~D Rivi{\`e}re and Lara~M Rangel.
\newblock Spike-field coherence and firing rate profiles of ca1 interneurons during an associative memory task.
\newblock In \emph{Association for Women in Mathematics Research Symposium}, pages 161--171. Springer, 2017.

\bibitem[Rivi{\`e}re et~al.(2022)Rivi{\`e}re, Schamberg, Coleman, and Rangel]{riviere2022modeling}
Pamela~D Rivi{\`e}re, Gabriel Schamberg, Todd~P Coleman, and Lara~M Rangel.
\newblock Modeling relationships between rhythmic processes and neuronal spike timing.
\newblock \emph{Journal of Neurophysiology}, 128\penalty0 (3):\penalty0 593--610, 2022.

\bibitem[Rivi{\`e}re et~al.(2024)Rivi{\`e}re, Beatty-Mart{\'\i}nez, and Trott]{riviere2024evaluating}
Pamela~D Rivi{\`e}re, Anne~L Beatty-Mart{\'\i}nez, and Sean Trott.
\newblock Evaluating contextualized representations of (spanish) ambiguous words: A new lexical resource and empirical analysis.
\newblock \emph{arXiv preprint arXiv:2406.14678}, 2024.

\bibitem[Rivière and Trott(2025)]{riviere2025tacl}
Pamela Rivière and Sean Trott.
\newblock Start making sense(s): A developmental probe of attention specialization using lexical ambiguity.
\newblock \emph{arXiv preprint}, 2025.

\bibitem[Saxon et~al.(2024)Saxon, Holtzman, West, Wang, and Saphra]{saxon2024benchmarks}
Michael Saxon, Ari Holtzman, Peter West, William~Yang Wang, and Naomi Saphra.
\newblock Benchmarks as microscopes: A call for model metrology.
\newblock In \emph{First Conference on Language Modeling}, 2024.
\newblock URL \url{https://openreview.net/forum?id=bttKwCZDkm}.

\bibitem[Schrimpf et~al.(2021)Schrimpf, Blank, Tuckute, Kauf, Hosseini, Kanwisher, Tenenbaum, and Fedorenko]{schrimpf2021neural}
Martin Schrimpf, Idan~Asher Blank, Greta Tuckute, Carina Kauf, Eghbal~A Hosseini, Nancy Kanwisher, Joshua~B Tenenbaum, and Evelina Fedorenko.
\newblock The neural architecture of language: Integrative modeling converges on predictive processing.
\newblock \emph{Proceedings of the National Academy of Sciences}, 118\penalty0 (45):\penalty0 e2105646118, 2021.

\bibitem[Singh et~al.(2024)Singh, Moskovitz, Hill, Chan, and Saxe]{10.5555/3692070.3693925}
Aaditya~K. Singh, Ted Moskovitz, Felix Hill, Stephanie C.~Y. Chan, and Andrew~M. Saxe.
\newblock What needs to go right for an induction head? a mechanistic study of in-context learning circuits and their formation.
\newblock In \emph{Proceedings of the 41st International Conference on Machine Learning}, ICML'24. JMLR.org, 2024.

\bibitem[Tigges et~al.(2024)Tigges, Hanna, Yu, and Biderman]{tigges2024llm}
Curt Tigges, Michael Hanna, Qinan Yu, and Stella Biderman.
\newblock Llm circuit analyses are consistent across training and scale.
\newblock \emph{arXiv preprint arXiv:2407.10827}, 2024.

\bibitem[van~der Wal et~al.(2025)van~der Wal, Lesci, M{\"u}ller-Eberstein, Saphra, Schoelkopf, Zuidema, and Biderman]{van2025polypythias}
Oskar van~der Wal, Pietro Lesci, Max M{\"u}ller-Eberstein, Naomi Saphra, Hailey Schoelkopf, Willem Zuidema, and Stella Biderman.
\newblock Polypythias: Stability and outliers across fifty language model pre-training runs.
\newblock In \emph{Proceedings of the Thirteenth International Conference on Learning Representations (ICLR 2025)}, pages 1--25, 2025.

\bibitem[Wang et~al.(2022)Wang, Variengien, Conmy, Shlegeris, and Steinhardt]{wang2022interpretability}
Kevin Wang, Alexandre Variengien, Arthur Conmy, Buck Shlegeris, and Jacob Steinhardt.
\newblock Interpretability in the wild: a circuit for indirect object identification in gpt-2 small.
\newblock \emph{arXiv preprint arXiv:2211.00593}, 2022.

\bibitem[Wolf et~al.(2020)Wolf, Debut, Sanh, Chaumond, Delangue, Moi, Cistac, Rault, Louf, Funtowicz, Davison, Shleifer, von Platen, Ma, Jernite, Plu, Xu, Scao, Gugger, Drame, Lhoest, and Rush]{wolf-etal-2020-transformers}
Thomas Wolf, Lysandre Debut, Victor Sanh, Julien Chaumond, Clement Delangue, Anthony Moi, Pierric Cistac, Tim Rault, Rémi Louf, Morgan Funtowicz, Joe Davison, Sam Shleifer, Patrick von Platen, Clara Ma, Yacine Jernite, Julien Plu, Canwen Xu, Teven~Le Scao, Sylvain Gugger, Mariama Drame, Quentin Lhoest, and Alexander~M. Rush.
\newblock Transformers: State-of-the-art natural language processing.
\newblock In \emph{Proceedings of the 2020 Conference on Empirical Methods in Natural Language Processing: System Demonstrations}, pages 38--45, Online, October 2020. Association for Computational Linguistics.
\newblock URL \url{https://www.aclweb.org/anthology/2020.emnlp-demos.6}.

\bibitem[Yax et~al.(2024)Yax, Oudeyer, and Palminteri]{yax2024phylolm}
Nicolas Yax, Pierre-Yves Oudeyer, and Stefano Palminteri.
\newblock Phylolm: Inferring the phylogeny of large language models and predicting their performances in benchmarks.
\newblock \emph{arXiv preprint arXiv:2404.04671}, 2024.

\bibitem[Yin and Steinhardt(2025)]{yin2025which}
Kayo Yin and Jacob Steinhardt.
\newblock Which attention heads matter for in-context learning?
\newblock In \emph{Forty-second International Conference on Machine Learning}, 2025.
\newblock URL \url{https://openreview.net/forum?id=C7XmEByCFv}.

\bibitem[Zhang et~al.(2024)Zhang, Yu, Zang, Eickhoff, and Pavlick]{zhang2024same}
Ruochen Zhang, Qinan Yu, Matianyu Zang, Carsten Eickhoff, and Ellie Pavlick.
\newblock The same but different: Structural similarities and differences in multilingual language modeling.
\newblock \emph{arXiv preprint arXiv:2410.09223}, 2024.

\bibitem[Zhang et~al.(2025)Zhang, Yu, Zang, Eickhoff, and Pavlick]{zhang2025the}
Ruochen Zhang, Qinan Yu, Matianyu Zang, Carsten Eickhoff, and Ellie Pavlick.
\newblock The same but different: Structural similarities and differences in multilingual language modeling.
\newblock In \emph{The Thirteenth International Conference on Learning Representations}, 2025.
\newblock URL \url{https://openreview.net/forum?id=NCrFA7dq8T}.

\bibitem[Zhao et~al.(2025)Zhao, Qin, Alvarez-Melis, Kakade, and Saphra]{zhao2025distributional}
Rosie Zhao, Tian Qin, David Alvarez-Melis, Sham Kakade, and Naomi Saphra.
\newblock Distributional scaling laws for emergent capabilities.
\newblock \emph{arXiv preprint arXiv:2502.17356}, 2025.

\end{thebibliography}

\clearpage
\appendix

\section{Individual head trajectories in 14m}\label{appendix:ind_heads}

\begin{figure}
    \centering
    \begin{subfigure}[t]{0.45\textwidth}
        \centering
        \includegraphics[width=\linewidth]{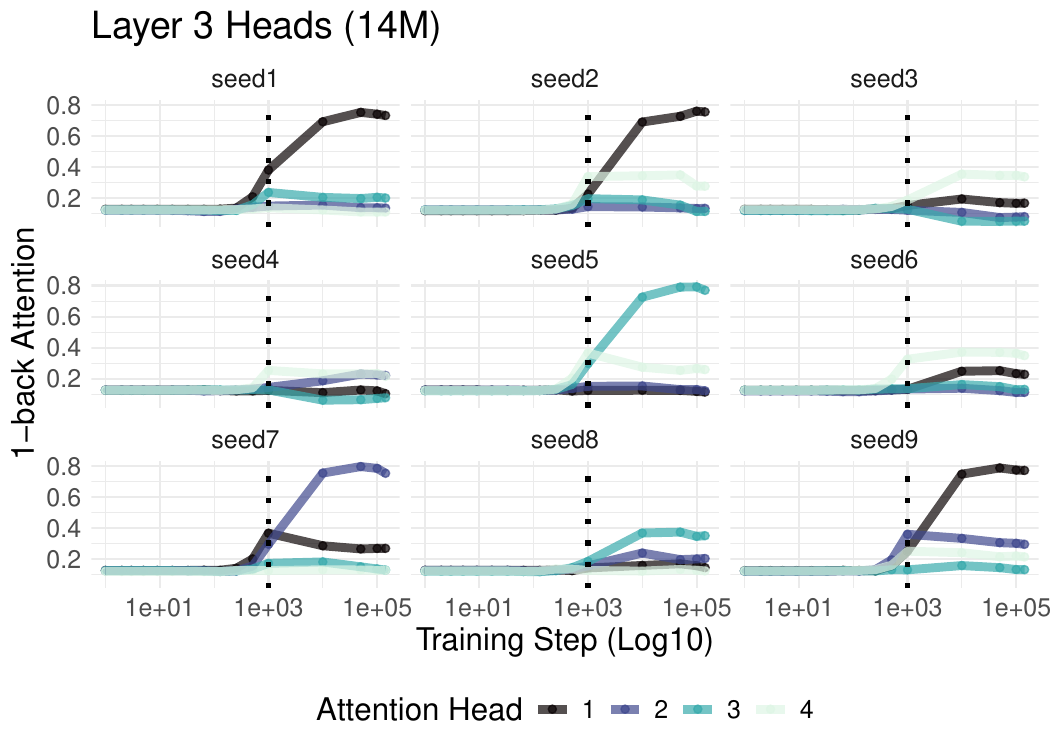}
        \caption{1-back attention for layer 3 heads across Pythia-14M seeds and pre-training checkpoints.}
        \label{supp_fig:layer3}
    \end{subfigure}
    \hfill
    \begin{subfigure}[t]{0.45\textwidth}
        \centering
        \includegraphics[width=\linewidth]{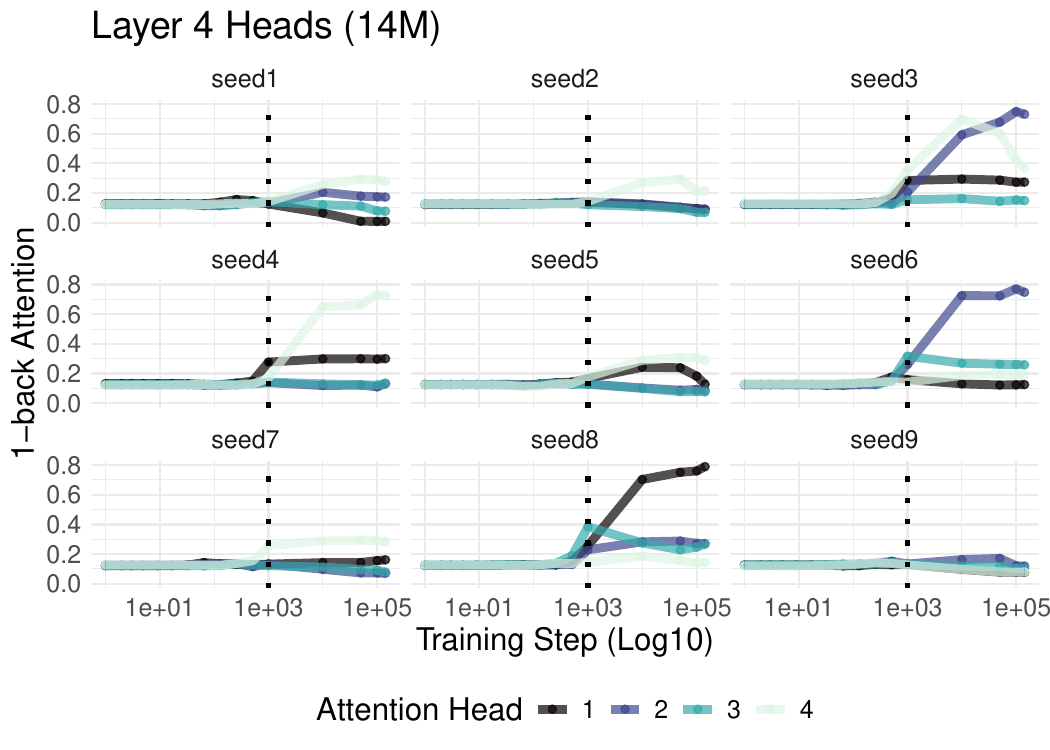}
        \caption{1-back attention for layer 4 heads across Pythia-14M seeds and pre-training checkpoints.}
        \label{supp_fig:layer4}
    \end{subfigure}
    \caption{In each random seed of Pythia-14M, 1-back attention heads emerged at roughly similar training checkpoints.}
    \label{supp_fig:ind_heads}
\end{figure}

The developmental trajectories depicted in the primary manuscript collapsed across heads and layers for the purpose of illustrating temporal patterns across all models tested. Here, I depict the trajectories of 1-back attention for each individual head in each random seed of Pythia-14M for layer 3 (Figure \ref{supp_fig:layer3}) and layer 4 (Figure \ref{supp_fig:layer4}).

\section{Consistency across stories}\label{appendix:stories}

One question that arises is about the generalizability of the results reported in the primary manuscript across \textit{input sentences}. That is, how dependent is the developmental trajectory observed upon the corpus used to assess attention head function? To address this, I plotted the maximum 1-back attention for each random seed of Pythia-14M for each \textit{story} from the Natural Stories Corpus. The developmental trajectories are extremely similar, with changes in attention beginning around $10^3$ training steps for all random seeds, for all stories.

\section{Multi-dimensional scaling of seeds}\label{appendix:mds}

\begin{figure}
    \centering
    \includegraphics[width=\linewidth]{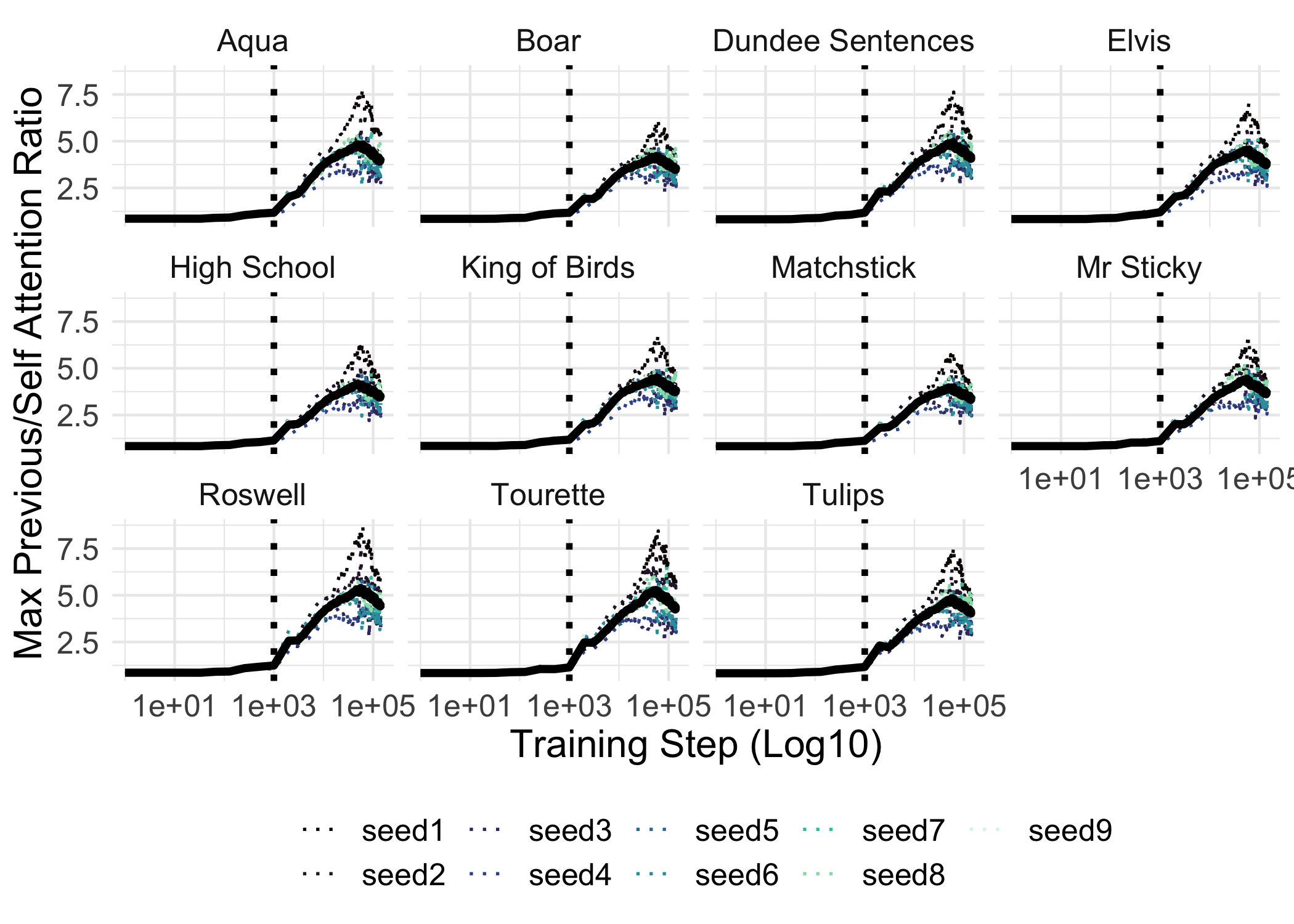}
    \caption{Developmental trajectory of putative 1-back attention heads across random seeds of Pythia-14M and across the story corpora used to assess this behavior. Across all heads at a given checkpoint for a given model, the maximum 1-back attention was calculated.}

    \label{supp_fig:multi-panel_layers}
\end{figure}

In the primary manuscript, I reported the average correlation between seeds belonging to the same architecture and between seeds belonging to different architectures; a key finding was among seeds of \textit{different architectures}, stronger temporal convergences were observed when both instances being compared were larger. 

\begin{figure}
    \centering
    \includegraphics[width=.9\linewidth]{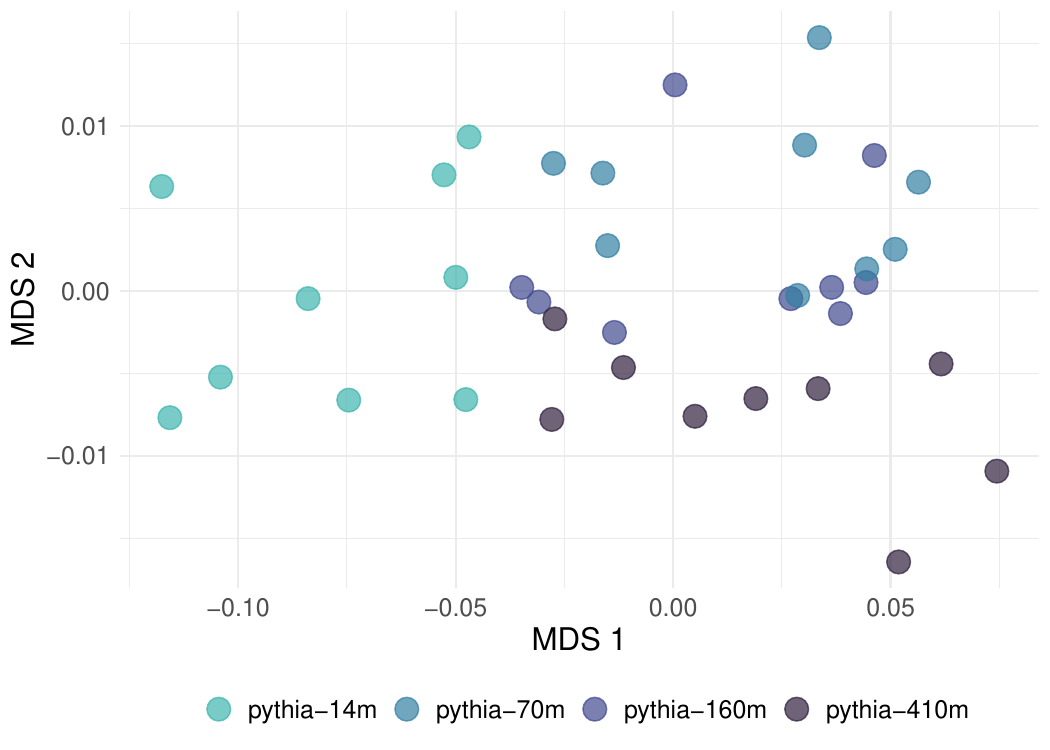}
    \caption{Results of applying multi-dimensional scaling (MDS) to the correlation matrix of all model instance pairs. Each MDS point represents a given seed of a particular model.}
    \label{fig:mds}
\end{figure}

Another perspective on this result comes from embedding the correlation matrix of all model instances in a $2D$ space using multi-dimensional scaling (MDS) \citep{carroll1998multidimensional}. As depicted in Figure \ref{fig:mds}, the first MDS component appeared to track model size. Moreover, seeds of smaller models ($14M$) exhibited tight clustering and relatively larger separation from larger models. 

I quantified this trend by first calculating the centroid among each set of seeds (i.e., for 14M, 70M, etc.). I then calculated the Separation Ratio, defined as the mean distance of each centroid to other centroids, divided by the mean distance of each point within a model class to the centroid of that class. It decreases systematically with model size, reflecting the combination of wider internal spread and closer inter-cluster proximity in larger models. The Separation Ratio systematically declined with model size: 14M was the largest ($11.4$), followed by 70M ($3.37$), and 160M ($2.42$). The results are nuanced, however, as 410M exhibited a larger Separation Ratio than 70M and 160M ($4.0$), though still considerably smaller than 14M; see also Figure \ref{fig:fig1}.

Together, this suggests that at least along the temporal axis, smaller model instances of the same architecture are relatively different from other models---conversely, larger models of different architectures are relatively more similar to each other.

\newpage
\section*{NeurIPS Paper Checklist}

\begin{enumerate}

\item {\bf Claims}
    \item[] Question: Do the main claims made in the abstract and introduction accurately reflect the paper's contributions and scope?
    \item[] Answer: \answerYes{} 
    \item[] Justification: The goal of the paper is to argue that generalizability is a major epistemological challenge in the study of LLMs, propose a theoretical framework for investigating it, and present the results of a novel case study; the abstract and introduction summarizes this.

\item {\bf Limitations}
    \item[] Question: Does the paper discuss the limitations of the work performed by the authors?
    \item[] Answer: \answerYes{} 
    \item[] Justification: Yes, there is an Objections and Limitations section devoted to identifying limitations, particularly in the theoretical framework proposed (which is the central point of the paper). 

\item {\bf Theory assumptions and proofs}
    \item[] Question: For each theoretical result, does the paper provide the full set of assumptions and a complete (and correct) proof?
    \item[] Answer: \answerNA{} 
    \item[] Justification: The results in the paper are empirical results and I have attempted to motivate and contextualize them (including the assumptions behind various operationalizations, etc.); there are no theoretical results depending on a proof.

    \item {\bf Experimental result reproducibility}
    \item[] Question: Does the paper fully disclose all the information needed to reproduce the main experimental results of the paper to the extent that it affects the main claims and/or conclusions of the paper (regardless of whether the code and data are provided or not)?
    \item[] Answer: \answerYes{} 
    \item[] Justification: The Methods section discloses the models run and data used as input to the models; the Results section describes the details of specific operationalizations and statistical analyses. Additionally, a link to a GitHub repository with code and data required to reproduce these analysis can be found at \url{https://github.com/seantrott/mechinterp_generalizability}. 

\item {\bf Open access to data and code}
    \item[] Question: Does the paper provide open access to the data and code, with sufficient instructions to faithfully reproduce the main experimental results, as described in supplemental material?
    \item[] Answer: \answerYes{} 
    \item[] Justification: A link to a GitHub repository with code and data required to reproduce these analysis can be found at \url{https://github.com/seantrott/mechinterp_generalizability}. 

\item {\bf Experimental setting/details}
    \item[] Question: Does the paper specify all the training and test details (e.g., data splits, hyperparameters, how they were chosen, type of optimizer, etc.) necessary to understand the results?
    \item[] Answer: \answerNA{} 
    \item[] Justification: No model training was performed.

\item {\bf Experiment statistical significance}
    \item[] Question: Does the paper report error bars suitably and correctly defined or other appropriate information about the statistical significance of the experiments?
    \item[] Answer: \answerYes{} 
    \item[] Justification: Statistical results are reported where relevant, including with standard errors and significance. Additionally, figures displaying means also include confidence intervals reflecting the standard error; other figures display the ``raw'' data.

\item {\bf Experiments compute resources}
    \item[] Question: For each experiment, does the paper provide sufficient information on the computer resources (type of compute workers, memory, time of execution) needed to reproduce the experiments?
    \item[] Answer: \answerYes{} 
    \item[] Justification: The Methods section includes a sentence describing the compute environment in which models were run.
    
\item {\bf Code of ethics}
    \item[] Question: Does the research conducted in the paper conform, in every respect, with the NeurIPS Code of Ethics \url{https://neurips.cc/public/EthicsGuidelines}?
    \item[] Answer: \answerYes{} 
    \item[] Justification: Human-subjects data was not used, and all corpus data presented to the model was extracted from a publicly available research corpus.

\item {\bf Broader impacts}
    \item[] Question: Does the paper discuss both potential positive societal impacts and negative societal impacts of the work performed?
    \item[] Answer: \answerNo{} 
    \item[] Justification: The answer is ``yes'' if this includes the impact on the scientific community, given that the goal of the work is to shape the research conversation around mechanistic interpretability. But it is ``no'' otherwise, as the paper does not discuss broader societal implications at length.
    
\item {\bf Safeguards}
    \item[] Question: Does the paper describe safeguards that have been put in place for responsible release of data or models that have a high risk for misuse (e.g., pretrained language models, image generators, or scraped datasets)?
    \item[] Answer: \answerNo{} 
    \item[] Justification: No new corpus data or language models were created in this research; the data outputs reflect the behavior of specific model components (attention heads) and are unlikely candidates for misuse.

\item {\bf Licenses for existing assets}
    \item[] Question: Are the creators or original owners of assets (e.g., code, data, models), used in the paper, properly credited and are the license and terms of use explicitly mentioned and properly respected?
    \item[] Answer: \answerYes{} 
    \item[] Justification: The language models and corpus data used in the paper are cited, and the licenses for each are also described.

\item {\bf New assets}
    \item[] Question: Are new assets introduced in the paper well documented and is the documentation provided alongside the assets?
    \item[] Answer: \answerYes{} 
    \item[] Justification: All data and code is accompanied by documentation (though new assets are minimal).

\item {\bf Crowdsourcing and research with human subjects}
    \item[] Question: For crowdsourcing experiments and research with human subjects, does the paper include the full text of instructions given to participants and screenshots, if applicable, as well as details about compensation (if any)? 
    \item[] Answer: \answerNA{} 
    \item[] Justification: The paper does not involve crowdsourcing nor research with human subjects. 

\item {\bf Institutional review board (IRB) approvals or equivalent for research with human subjects}
    \item[] Question: Does the paper describe potential risks incurred by study participants, whether such risks were disclosed to the subjects, and whether Institutional Review Board (IRB) approvals (or an equivalent approval/review based on the requirements of your country or institution) were obtained?
    \item[] Answer: \answerNA{} 
    \item[] Justification: The paper does not involve crowdsourcing nor research with human subjects.

\item {\bf Declaration of LLM usage}
    \item[] Question: Does the paper describe the usage of LLMs if it is an important, original, or non-standard component of the core methods in this research? Note that if the LLM is used only for writing, editing, or formatting purposes and does not impact the core methodology, scientific rigorousness, or originality of the research, declaration is not required.
    \item[] Answer: \answerNA{} 
    \item[] Justification: The \textit{focus} of the research is LLMs, but LLMs were not used in this research. 

\end{enumerate}

\end{document}